\documentclass[]{agile}
\usepackage{float}
\usepackage[dvipsnames]{xcolor}
\usepackage{graphicx}
\usepackage{ulem}
\usepackage{color}
\usepackage{parrun}
\usepackage{fancyhdr}
\usepackage{framed, color}
\usepackage{lipsum}
\usepackage{tcolorbox}
\usepackage{sectsty}
\usepackage{enumitem}
\usepackage{etoolbox}
\usepackage{titlesec}
\usepackage{hyperref}
\hypersetup{
    colorlinks=true,
    filecolor=magenta,      
    urlcolor=blue,
    linkcolor=blue,
    citecolor=black,
    }
\usepackage[a4paper]{geometry}
 \newgeometry{left=4.8cm, right=0.2cm, headheight=-0.2cm, top=2.3cm, bottom=1.3cm}
\usepackage{caption}
\usepackage{subcaption}

\fancypagestyle{firststyle}
{
   \fancyhf{}
  \rhead{\raisebox{-1.2\height}{\includegraphics[width=7.44cm]{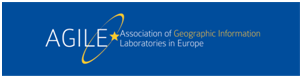}}}
 
}

\fancypagestyle{secondstyle}
{
   \fancyhf{}
  \chead[\raisebox{-1.2\height}{}]{\raisebox{-1.2\height}{}}
  \newgeometry{left=4.8cm, right=0.2cm, top=4cm}
  }
 
\definecolor{AGILEblue}{RGB}{0, 77, 156}

\titlespacing\section{0pt}{24pt plus 0pt minus 2pt}{12pt plus 0pt minus 2pt}
\titlespacing\subsection{0pt}{12pt plus 0pt minus 2pt}{12pt plus 0pt minus 2pt}
\sectionfont{\large}
\subsectionfont{\normalsize}
\newcommand{\revise}{\textcolor{black}}

\begin{document}
%Insert your title here
\title{\textcolor{AGILEblue}{Geographic Question Answering: Challenges, Uniqueness, Classification, and Future Directions}}

\runningtitle{Geographic Question Answering}

% \author[]{Anonymized Author}
% \runningauthor{Gengchen Mai et al.$^a$}

\author{\normalfont \large Gengchen Mai\textsuperscript{a,b}(corresponding author), Krzysztof Janowicz\textsuperscript{a,b}, Rui Zhu\textsuperscript{a,b}, Ling Cai\textsuperscript{a,b}, and Ni Lao%\textsuperscript{c}  
\newline
\newline \normalsize \color{blue} \uline{gengchen\_mai@ucsb.edu}, 
\uline{janowicz@ucsb.edu}, 
\uline{ruizhu@ucsb.edu}, 
\uline{lingcai@ucsb.edu}, 
\uline{noon99@gmail.com}  
\newline 
\newline \color{black} \textsuperscript{a}STKO Lab, Department of Geography, University of California, Santa Barbara, CA, USA
\newline \textsuperscript{b}Center for Spatial Studies, University of California, Santa Barbara, CA, USA
% \newline 
% \textsuperscript{c}SayMosaic Inc., Mountain View, CA, USA
} 
\runningauthor{Gengchen Mai et al.$^a$}

%\newline \textsuperscript{b}Department example, University example, city, country
%\author{Anonymous}
%\correspondence{NAME (EMAIL)}

% \author[]{Anonymized Author}
% \author{\normalfont \large Author One\textsuperscript{a}(corresponding author), Author Two\textsuperscript{a} and Author Three\textsuperscript{b} \newline
% \newline \normalsize \color{blue} \uline{emailaddress@author.one}, \uline{emailaddress@author.two},  \uline{emailaddress@author.three} \newline 
% \newline \color{black} \textsuperscript{a}Department example, University example, city, country \newline \textsuperscript{b}Department example, University example, city, country} 

% \correspondence{Gengchen Mai (gengchen\_mai@ucsb.edu)}

% \runningauthor{Author One$^a$(corresponding author), Author Two$^a$ and Author Three$^b$}
% \runningauthor{Gengchen Mai et al.$^a$}

\maketitle
\firstpage{1}

%% logo header on the other pages of the article
%\thispagestyle{firststyle}
%\renewcommand{\familydefault}{\rmdefault}
%\newgeometry{left=2cm,right=2cm,top=2.5cm,bottom=2.8cm}
\nolinenumbers
\def\labelitemi{$\bullet$}
\setlength{\parskip}{4pt}
\setlength{\parindent}{0pt}

\begin{abstract}
As an important part of Artificial Intelligence (AI), Question Answering (QA) aims at generating answers to questions phrased in natural language. While there has been substantial progress in open-domain question answering, QA systems are still struggling to answer questions which involve geographic entities or concepts and that require spatial operations.  In this paper, we discuss the problem of geographic question answering (GeoQA). We first investigate the reasons why geographic questions are difficult to answer by analyzing challenges of geographic questions. We discuss the uniqueness of geographic questions compared to general QA. Then we review existing work on GeoQA and classify them by the types of questions they can address. Based on this survey, we provide a generic classification framework for geographic questions. Finally, we conclude our work by pointing out unique future research directions for GeoQA.

\end{abstract}
\section{Introduction}  \label{sec:intro}

\textit{
% I don’t think Google Translate, which involves machine translation, is the only language problem. 
``Another example of a good language problem is question answering, like “What’s the second-biggest city in California that is not near a river?” If I typed that sentence into Google currently, I’m not likely to get a useful response.''}\footnote{Interestingly, now Google can correctly answer this geographic question based on reading comprehension over an Wikipedia article. 
% But we guess Google was not able to handle it back to 2014 since that article was not available at that time.
Nevertheless, using reading comprehension to answer this kind of geographic questions is problematic and suffers from data sparsity issue (See Section \ref{sec:geoqa-problem}).}
-- Dr. Michael Jordan, UC Berkeley~\citep{jordan2014}
%https://spectrum.ieee.org/artificial-intelligence/machine-learning/machinelearning-maestro-michael-jordan-on-the-delusions-of-big-data-and-other-huge-engineering-efforts

Question Answering (QA) lies at the intersection of natural language processing (NLP), information retrieval (IR), knowledge representation, and computational linguistics. It aims at generating or retrieving answers to questions asked in natural language~\citep{mishra2016survey}. Question answering is an important part of artificial intelligence (AI) research~\citep{turing1950} and has recently permeated to our daily lives. Many commercial language understanding systems or voice control systems are widely adopted by the general public such as Apple Siri, Amazon Alexa,  Google's assistant,  Xiaomi Xiaoai, and so on. 

Generally speaking, question answering systems can be classified into three categories based on the types of data sources~\citep{mishra2016survey}: unstructured data-based QA \citep{rajpurkar2016squad,miller2016key,yang2017learning,chen2017reading,mai2018poireviewqa}, semi-structured table-based QA \citep{pasupat2015compositional}, and structured data source-based QA (so-called semantic parsing) \citep{zelle1996learning,yih2016value,liang2017neural,berant2013semantic,liang2018memory,chen2020neural}. Thanks to the recent development of multiple open domain QA datasets such as HotpotQA~\citep{yang2019end}, SQuAD Open~\citep{chen2017reading}, and Natural Questions Open~\citep{kwiatkowski2019natural},  %which aim at searching answers from a large corpus instead of searching within a small paragraph, 
research on unstructured data-based QA has made substantial progress~\citep{asai2020learning,karpukhin2020dense,xiong2020answering}.
Recently, we have also seen remarkable advancements in hybrid QA models which rely on different data sources, such as hybrid QA models based on both knowledge graphs and unstructured texts~\citep{sun2018open,xiong2019improving}.

\begin{figure*}
    \centering
    \begin{subfigure}[b]{0.48\textwidth}
        \centering
        \includegraphics[width=\textwidth]{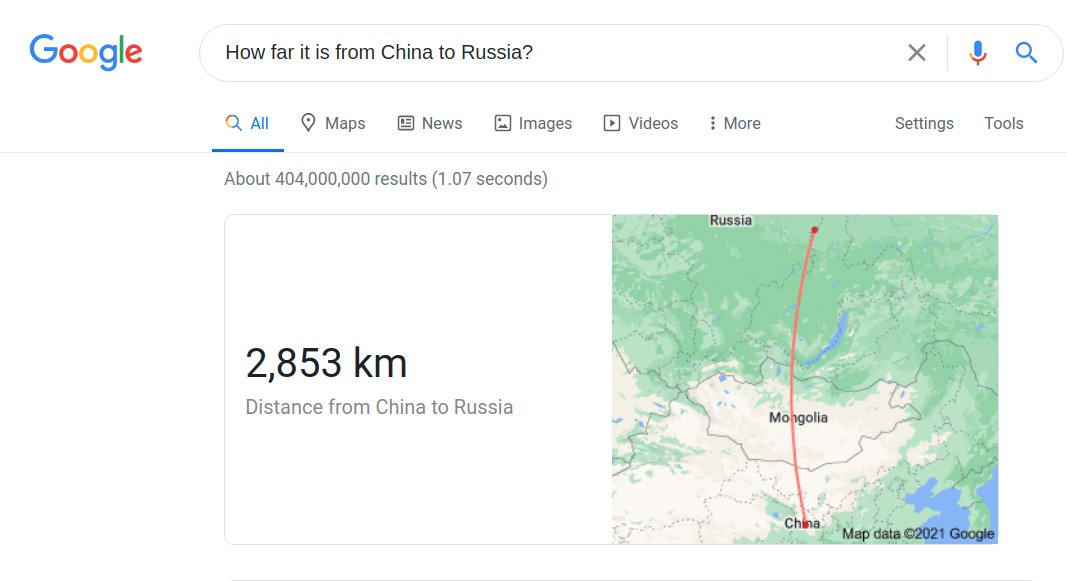}
        \caption[]%
		{Question A1: spatial proximity
		}
        \label{fig:qa1-1}
    \end{subfigure}
    \hfill
    \begin{subfigure}[b]{0.48\textwidth}
        \centering
        \includegraphics[width=\textwidth]{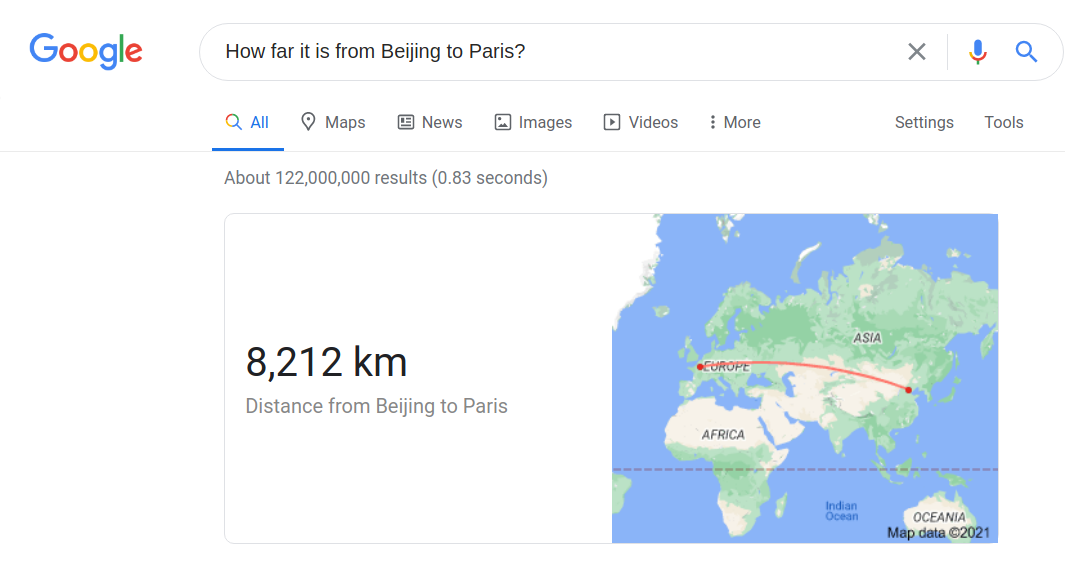}
        \caption[]%
		{Question A2: spatial proximity
		}
        \label{fig:qa1-2}
    \end{subfigure}
    \hfill
    \begin{subfigure}[b]{0.48\textwidth}
        \centering
        \includegraphics[width=\textwidth]{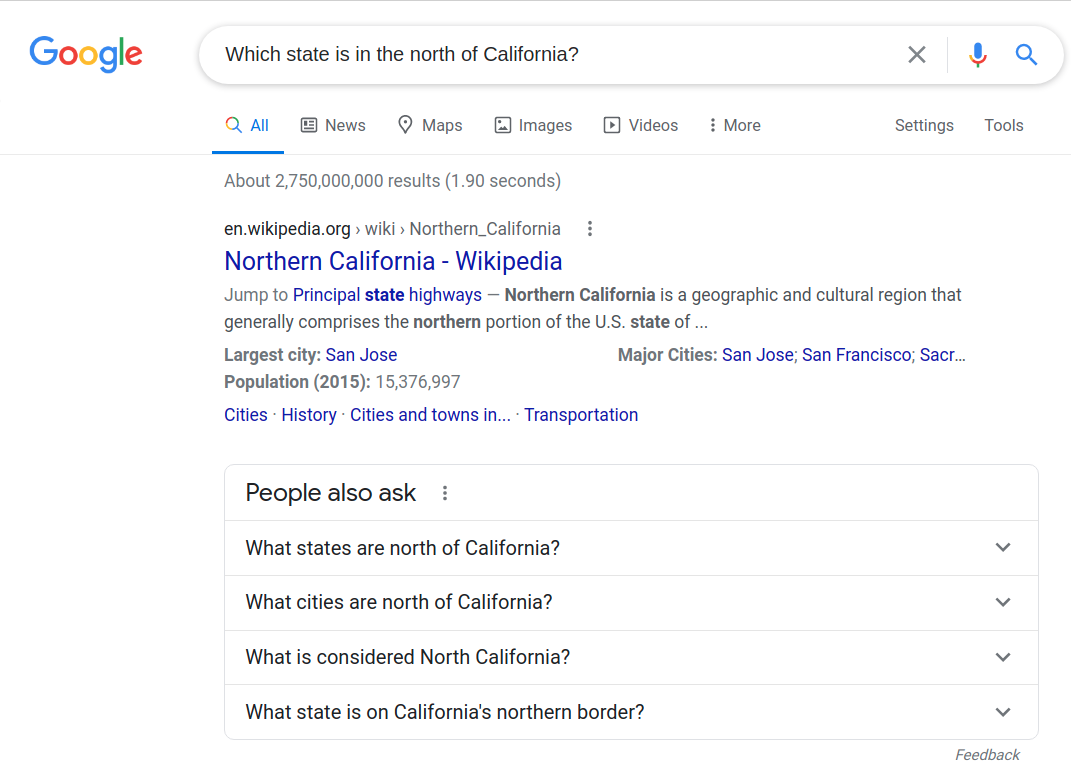}
        \caption[]%
		{Question B1: cardinal direction
		}
        \label{fig:qa2-1}
    \end{subfigure}
    \hfill
    \begin{subfigure}[b]{0.48\textwidth}
        \centering
        \includegraphics[width=\textwidth]{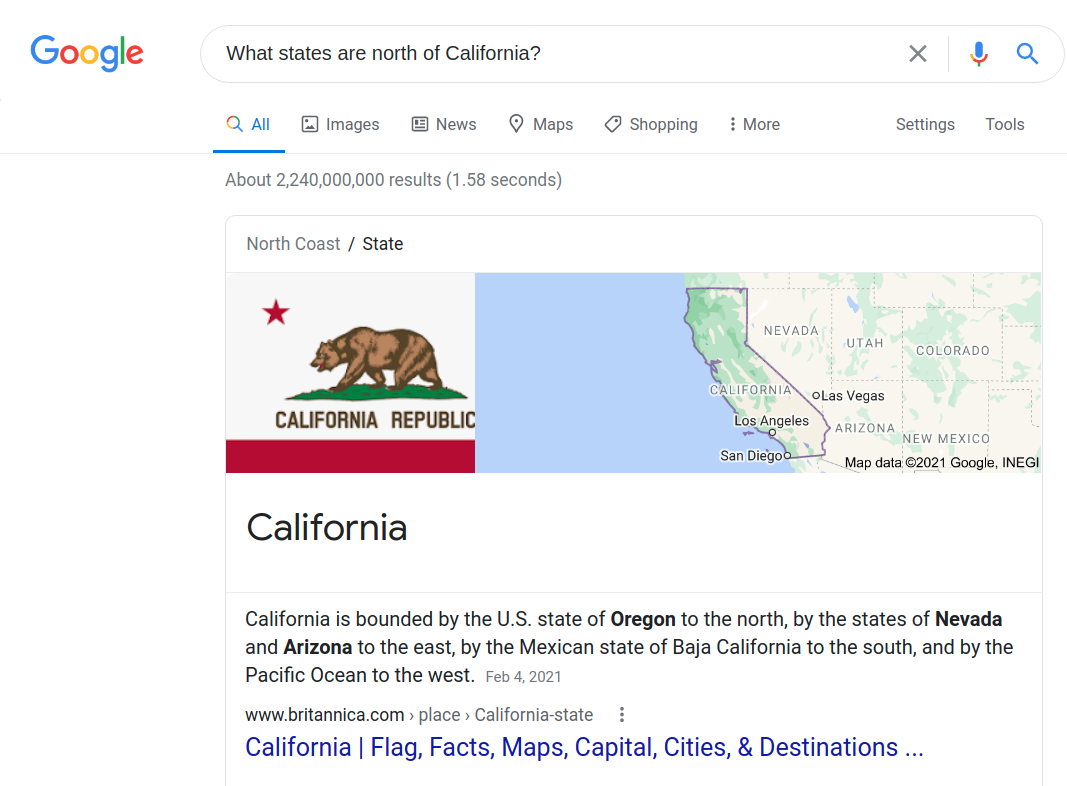}
        \caption[]%
		{Question B2: cardinal direction
		}
        \label{fig:qa2-2}
    \end{subfigure}
    \hfill
    \begin{subfigure}[b]{0.48\textwidth}
        \centering
        \includegraphics[width=\textwidth]{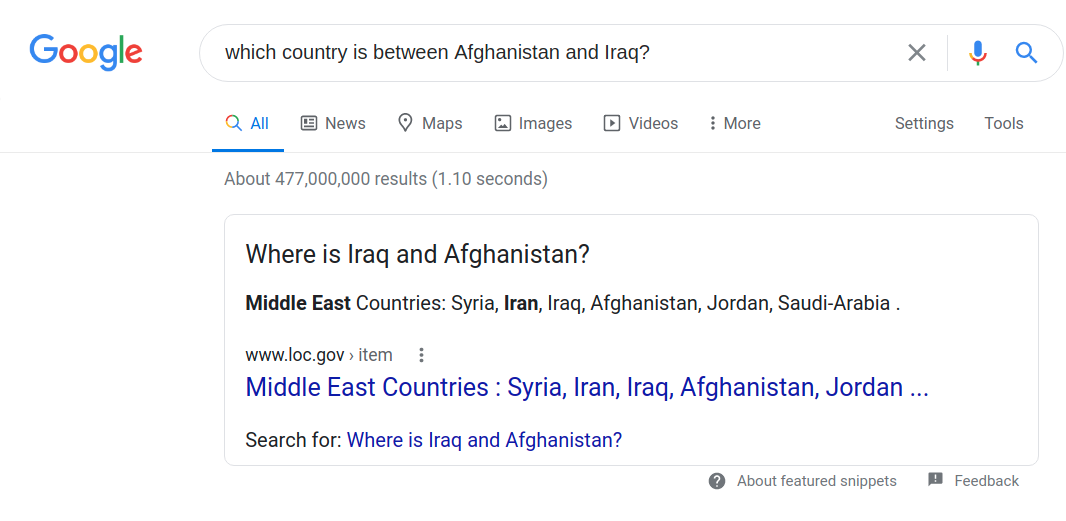}
        \caption[]%
		{Question C1: betweenness
		}
        \label{fig:qa3-1}
    \end{subfigure}
    \hfill
    \begin{subfigure}[b]{0.48\textwidth}
        \centering
        \includegraphics[width=\textwidth]{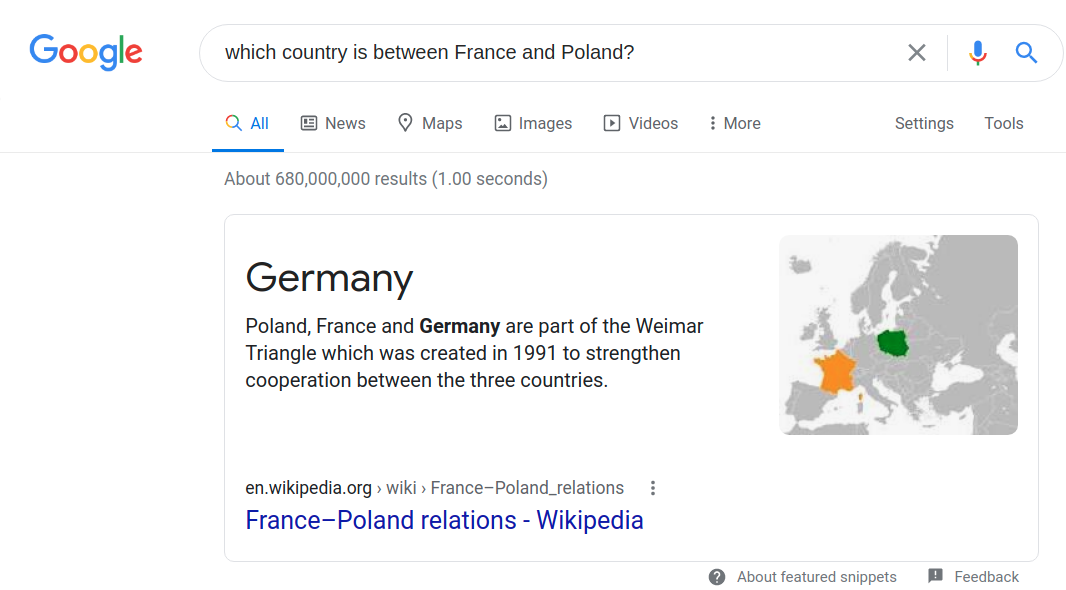}
        \caption[]%
		{Question C2: betweenness
		}
        \label{fig:qa3-2}
    \end{subfigure}
        \caption{Three pairs of geographic questions that \revise{show} the limitation of Google's question answering system. Google's question answering system fails to answer Question A1, B1, C1 while being able to handle A2, B2, C2 even if these three pairs A1 - A2, B1 - B2, C1 - C2 are quite similar to each other. All screenshots obtained on Feb. 17th, 2021.}
    \label{fig:challenge-geoqa}
\end{figure*}

Although the performance gap between human's and deep neural network-based QA models has been significantly reduced on reading comprehension style QA tasks~\citep{rajpurkar2016squad}, we still get a fairly poor performance when applying these models in the wild. Even commercial QA products such as Google question answering system are struggling to answer many \revise{simple} geographic questions. 
%One challenging geographic question is the one Dr. Michael Jordan mentioned during his interview with San Francisco writer Lee Gomes for IEEE Spectrum on 3 October 2014~\citep{jordan2014}. 
Figure \ref{fig:challenge-geoqa} shows several challenging geographic questions which shows the limitation of Google QA system that is powering their search.
% This problem gets worse when it comes to \textit{geographic questions}.

In this work, we define \textit{geographic questions} as questions that involve geographic entities (e.g., Los Angeles, Eastern Sierra), geographic concepts (e.g., feature types such as Building, City, State), or spatial relations (e.g. near to, north of, between) as parts of the natural language questions. 
Note that this definition is rather broad compared to related notions such as geo-analytical questions~\citep{scheider2020geo} which require geo-analytical workflows (in GIS) to answer them. 
The corresponding QA systems and processes are named geographic question answering (GeoQA). While some geographic questions are easy to answer such as \textit{what is the population of London} or \textit{where is Los Angeles} as they only require a simple property fact lookup in a knowledge base/graph, other geographic questions are more challenging to handle even for state-of-the-art (SOTA) question answering systems.
%such as Google question answering system. 

Figure \ref{fig:challenge-geoqa} shows three pairs of geographic questions which demonstrate the limitation of Google QA.
%the SOTA QA system. 
Question A1 \& A2, B1 \& B2, and C1 \& C2 involve three different types of spatial operations in order to answer geographic questions, namely spatial proximity, cardinal direction, and projective ternary relation (e.g., betweenness)~\citep{billen2004model}. While Google QA can provide meaningful answers to Question A2, B2, and C2 as shown in Figure \ref{fig:qa1-2}, \ref{fig:qa2-2}, and \ref{fig:qa3-2}, it can not handle simple variations of them (Question A1, B1, and C1 as shown in Figure \ref{fig:qa1-1}, \ref{fig:qa2-1}, and \ref{fig:qa3-1}). A1, A2, B1, and B2 are simple questions or so-called single-relation factoid questions~\citep{yin2016simple} which can be answered by using a single triple in a Knowledge Graph (KG), if available. C1 and C2 are expected to be answered based on two triples in a KG. These questions show interesting properties shared by geographic questions and give us hints about why geographic questions are difficult to handle.

%Research questions which this paper aims to answer are listed as follows: 
In this paper, we aim at answering the following three research questions:
\begin{enumerate} %[itemsep=-0.2em] 
	\item  \textit{Why} are geographic questions difficult to answer compared to generic questions?
% 	\item {What is the uniqueness of geographic questions and GeoQA?} 
	\item  {\textit{How} to classify geographic questions?}
	\item  {\textit{What} unique contributions can GIScience make in GeoQA in addition to SOTA approaches instead of reinventing the wheel?}
\end{enumerate}
%1) compared with other types of question, \textbf{why geographic questions are difficult to answer?} 2) \textbf{what is the uniqueness of geographic questions and GeoQA?} 3) \textbf{what is a meaningful classification of geographic questions?} 4) \textbf{as GIScientists, what unique contributions can we make in GeoQA to stand on the shoulders of giants and avoid reinventing the wheels?}

In the following, we will go through those geographic questions in Figure \ref{fig:challenge-geoqa} and discuss the reason why current QA system fail. 
Next, we discuss the uniqueness of geographic questions and GeoQA in Section \ref{sec:unique} from a conceptual level. 
Then, in Section \ref{sec:related}, we present  existing work on GeoQA by classifying them into different groups based on the types of questions they can handle and discuss pros and cons of \revise{them}. Section \ref{sec:geoqaclass} provides a detailed classification of geographic questions and \revise{discusses} the possible solutions and challenges of GeoQA for each question type. 
Last, we conclude this paper by discussing possible future research directions in GeoQA. % that GIScientists can 
%work on to 
%make a unique contribution.

% \pagestyle{secondstyle}
\section{Why Geographic Questions are Difficult to Answer?} \label{sec:geoqa-problem}

In this section, we discuss the reasons why geographic questions are hard to answer by using the three pairs of geographic questions presented in Figure \ref{fig:challenge-geoqa}. 
%and the failures are caused by different reasons.

%which will also provide us some concrete understanding of the uniqueness of geographic questions.

\begin{enumerate}
    \item \textit{QA systems usually lack proper spatial representations (i.e., points, polylines, or polygons) for geographic entities}. Question A1 shown in Figure \ref{fig:qa1-1} is actually a brain teaser question. The correct answer is 0 since China is adjacent to \revise{Russia} \citep{janowicz2015data}. Although Google QA successfully recognizes the geographic entities involved in the question -- China and \revise{Russia}, it picks the wrong spatial representation (i.e., points) for spatial proximity computation. In fact, it is common practice for many widely used knowledge graphs such as Wikidata and DBpedia to represent all geographic entities as points regardless of their scale. Consequently, many QA systems based on these KBs would inherit this limitation. 
    %The practices of computing distance between places with their centroids will also suffer from inconsistency in answers among different QA systems since they are based on different KBs.
    %\citet{janowicz2015data} pointed out similar issues in current QA systems as well.

    \item \textit{Polygon-based spatial operations, such as the calculation of spatial proximity and topological relations between geographic entities, are computationally expensive.} Many geographic entities are represented by polygons with thousands of vertices, and, thus, spatial operations performed on them are difficult to carry out on demand.
    For Question A1, although Google Maps has the polygon representations for China and \revise{Russia}, it seems to always pick point geometries for the sake of fast response time. 

    \item \textit{The selection of spatial operator is subject to context} - \textit{where} a user asks a question, \textit{when} they ask it, \textit{which} geographic entities they are comparing. Both Question A1 and A2 have exactly the same query template  - \textit{how far it is from X to Y}. The reason why Google QA can successfully answer Question A2 but not A1 is because the scales of the compared geographic entities are different. 
    For A2, Paris and Beijing are far enough and thus can be presented at a small map scale. Their fine-grained geometries, i.e., polygons, can be ``safely'' ignored and we can use points to represent their locations. However, as for \revise{Russia} and China in A1, since they are adjacent to each other, their polygon representations are too large to be ignored. 
    How to pick the correct spatial representations and their corresponding spatial operators is challenging and depends on the map scale tied to the question\footnote{\revise{Note that we assume all geometries in the underlying geospatial knowledge base of a GeoQA system share the same coordinate system such as WGS84. This is a common practice used by many geographic knowledge graphs and geospatial ontologies such as GeoSPARQL. If two geographic entities have different coordinate systems, we need to do coordinate system transformation before the QA process.}}.
    
    \item \textit{Reading comprehension based QA cannot easily handle geographic questions}. Instead of computing the answers based on the geometries of geographic entities, many SOTA QA systems try to answer geographic questions by answering questions based on text corpus \citep{karpukhin2020dense} %or querying a knowledge graph \citep{zelle1996learning} 
    which suffer from data sparsity. 
    For example, Google QA tries to answer cardinal direction questions such as Question B1, B2 in Figure \ref{fig:qa2-1}, \ref{fig:qa2-2} and projective ternary relation questions such as Question C1, C2 in Figure \ref{fig:qa3-1}, \ref{fig:qa3-2} by searching the answers from a text corpus (e.g., websites) instead of computing answers based on geometries. Sometimes text-corpus-based QA can work (Question B2, C2) if relevant information happens to exist in the corpus, but many times it fails (Question B1, C1). As for those binary spatial relation-based questions such as \textit{which city/county/state is in the north/south/east/west of X}, one cannot pre-compute all possible pairs of places for their cardinal direction relations since this leads to a combinatorial explosion. The situation gets even worse when we consider projective ternary spatial relations (e.g., betweenness) or n-ary spatial relations (e.g., surrounded by).
    
    \item \textit{It is difficult to identify the correct spatial relations given the large spatial language variability.} This can be clearly seen in Figure \ref{fig:qa2-1} in which ``north of California'' is misinterpreted as ``Northern California'' which in turn causes the QA failure. 
    In fact, the difficulty of recognizing spatial relations from natural language sentences has attracted a lot of \revise{attention} from the NLP and machine learning community \citep{kordjamshidi2020representation}, especially in the domain of visual question answering \citep{antol2015vqa}. Many papers are focusing on recognizing spatial relations which are viewpoint dependent \citep{ramalho2018encoding} such as \textit{on the left of this door}, \textit{on the right of this building}, \textit{behind this desk}. As for topological and cardinal direction relations, researchers still rely on \revise{rule-based} %or template-based 
    methods \citep{chen2014parameterized,punjani2018template}.
    
    \item \textit{Many spatial relations are conceptually vague and therefore difficult to represent computationally in structures like knowledge graphs and difficult to learn.} A typical example of vague spatial relations is \textit{near} \citep{worboys2001nearness,frank1992qualitative}. The search radius for the nearby geographic entities varies according to the map scale of the center entity. For example, Question \textit{Find restaurants near Marriott hotel} should use a smaller radius than Question \textit{Find small towns near London}. 
    Another example of vaguely defined spatial relations are cardinal directions (e.g., Question B1, B2) and ternary relations (e.g., Question C1, C2) between/among polygonal geographic entities. Is Nevada in the east or northeast of California? Moreover, the computation of cardinal directions between polygons is complex. 
    \citet{regalia2016volt} proposed a grid-point-based method which has $O(n^2)$ complexity \footnote{$n$ is the number of grid points in each polygon}.
    %\citet{regalia2016volt} proposed a grid-point-based method in which the direction between two polygons is decided based on a majority vote from the cardinal directions between all pairs of grid points inside each polygon.
    %They first obtained grids of points inside each polygon. Then they computed cardinal directions between every pair of grid points which come from the polygons respectively. A majority vote was used to decide the final cardinal direction between two polygons. 
    %This computation has $O(n^2)$ complexity \footnote{$n$ is the number of grid points in each polygon}. 
    As for Question B1 and B2 which search for all states north of California, this computation becomes prohibitively complex. Moreover, we cannot materialize all these cardinal direction relations in a KG beforehand either since this leads to a combinatorial explosion as we discussed above. Similarly, the betweenness relation among geographic entities is also vague and has high computation complexity. 
    %So an efficient operator to compute these spatial relations among polygonal geographic entities is in desperate need.
    
    \item \revise{\textit{
    There is a \textit{spurious program} issue mentioned by \citet{liang2017neural}}. A \textit{spurious program} is a program produced by a semantic parser which accidentally produces the correct answer but with the wrong QA logic, and thus does not generalize to other questions. 
    For example, when we ask for \textit{PlaceOfBirth} of a person, a spurious program may instead ask for \textit{PlaceOfDeath} while these two places are the same for this person. 
    Although a correct QA logic is vital, this kind of QA logic errors is hard to detect by the current standard QA evaluation protocol which is only based on answer comparison. In a weak supervision setting as \citet{liang2017neural} did, it is hard to distinguish \textit{spurious programs} from the correct program since the only QA annotations are the answers.
    Similarly, to improve the generalizability of a GeoQA system, it requires not only the correct answer but also the correct computational logic/spatial logic.
    %A correct answer to geographic question does not guarantee a correct workflow or QA logic which is important for a GeoQA system to generalize to other questions. 
    For example, although Google QA correctly answers Question C2 shown in Figure \ref{fig:qa3-2}, the answer ``Germany'' is extracted from a web page about the political and social cooperation of France, Poland and Germany, not a web page about the spatial configuration among these countries. Thus the logic used to answer this question is wrong and %This is just a ``lucky guess'' and 
    slightly changing the question may break the QA process. In other words, the generalizability of this QA model is low. The same issue exists in Question B2 as shown in Figure \ref{fig:qa2-2}. Although the correct answer ``Oregon'' is highlighted in the text snippet, several other incorrect answers are also highlighted such as ``Nevada'' and ``Arizona'', which also indicates an incorrect QA logic. 
    How to overcome the QA logic error and let the model really understand questions are interesting research directions for GeoQA and QA in general.
    }
\end{enumerate}

\subsection{Uncertainty and Vagueness of Geographic Information} \label{sec:geo-uncertain-vague}

One may further ask whether the problems shown in Figure \ref{fig:challenge-geoqa} would be alleviated if we had a GeoQA system which can successfully recognize the correct and efficient spatial relation/operator as well as the correct geographic entities and use their polygon geometries (if necessary) to compute the answer. The answer is still no because of the uncertainty of geometries \citep{regalia2017revisiting} and the vagueness of geographic concepts/entities \citep{bennett2002forest} which usually exists in real-world geographic datasets.

\subsubsection{Geometric Uncertainty} \label{sec:geo-uncertain}
Geometric uncertainty refers to the fact that the precise geometry of one geographic entity may vary according to the map scale, the data source, and map digitization process. According to the famous \textit{coastline paradox}\footnote{\url{https://en.wikipedia.org/wiki/Coastline\_paradox}}, \textit{the coastline of a landmass does not have a well-defined length}. %This counterintuitive observation can be explained by the fractal theory.
Uncertainty of geometries is in fact caused by the coastal paradox.
%Geometry uncertainty problem has a similar cause to the coastal paradox.
Because of the uncertainty, sometimes we cannot get the correct spatial relationships between/among geographic entities based on their (polygon) geometries which might be derived from one or several geographic datasets such as OpenStreetMap.

\begin{figure*}
    \centering
    \begin{subfigure}[b]{0.48\textwidth}
        \centering
        \includegraphics[width=\textwidth]{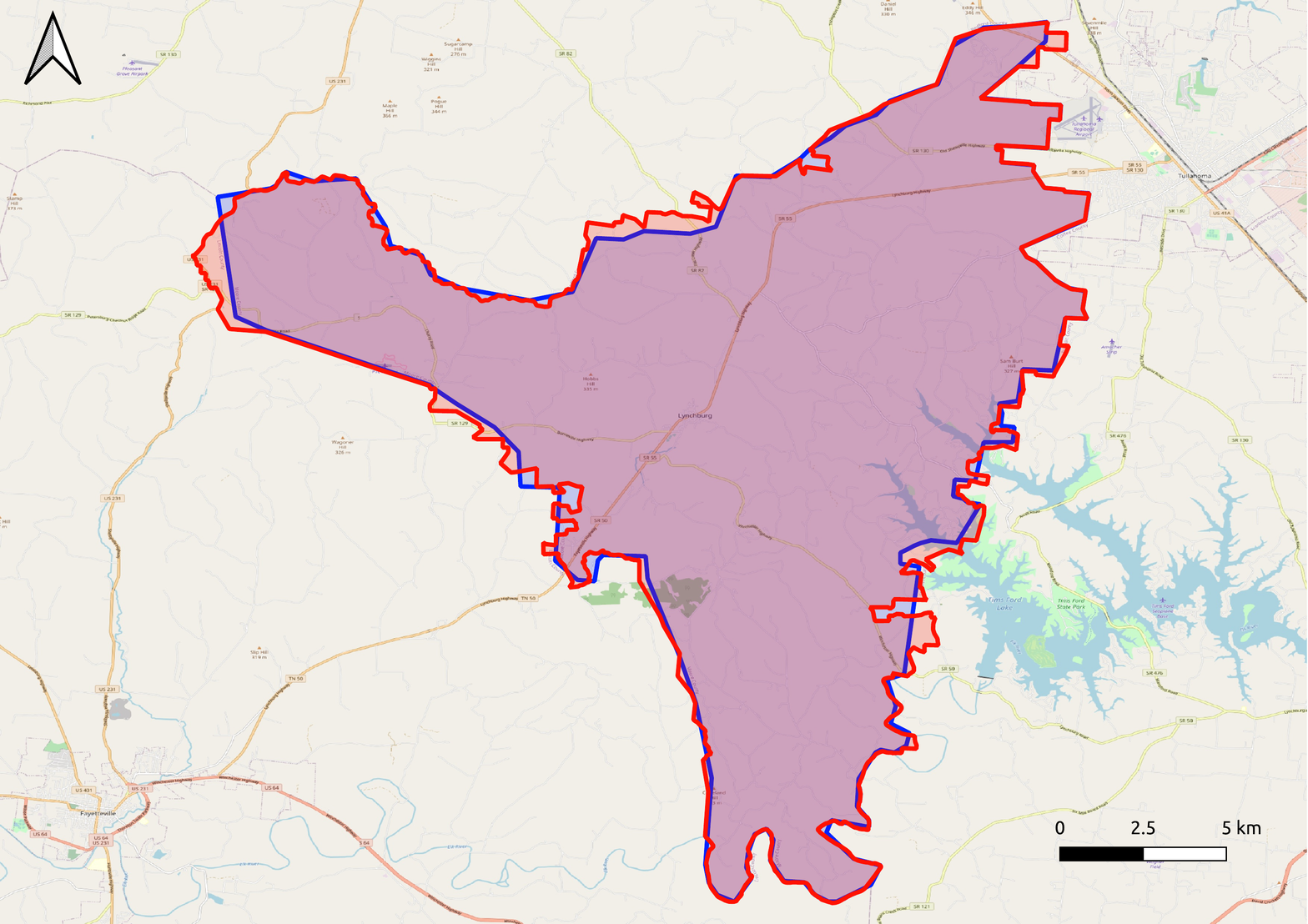}
        \caption[]%
		{Polygon Equivalent Error
		}
        \label{fig:equal}
    \end{subfigure}
    \hfill
    
    \begin{subfigure}[b]{0.48\textwidth}
        \centering
        \includegraphics[width=\textwidth]{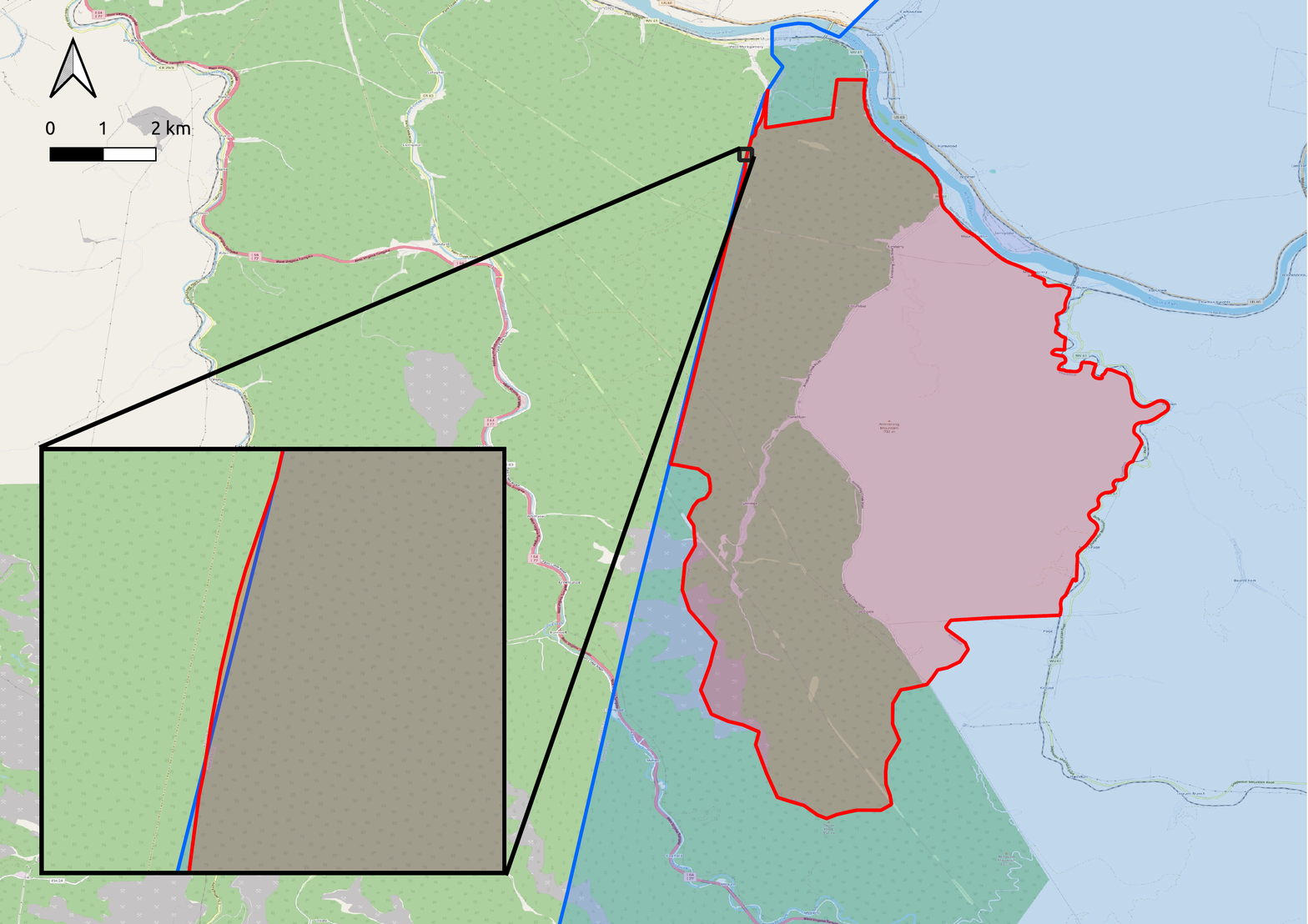}
        \caption[]%
		{Polygon Containment Error
		}
        \label{fig:ispartof}
    \end{subfigure}
    \hfill
    \begin{subfigure}[b]{0.48\textwidth}
        \centering
        \includegraphics[width=\textwidth]{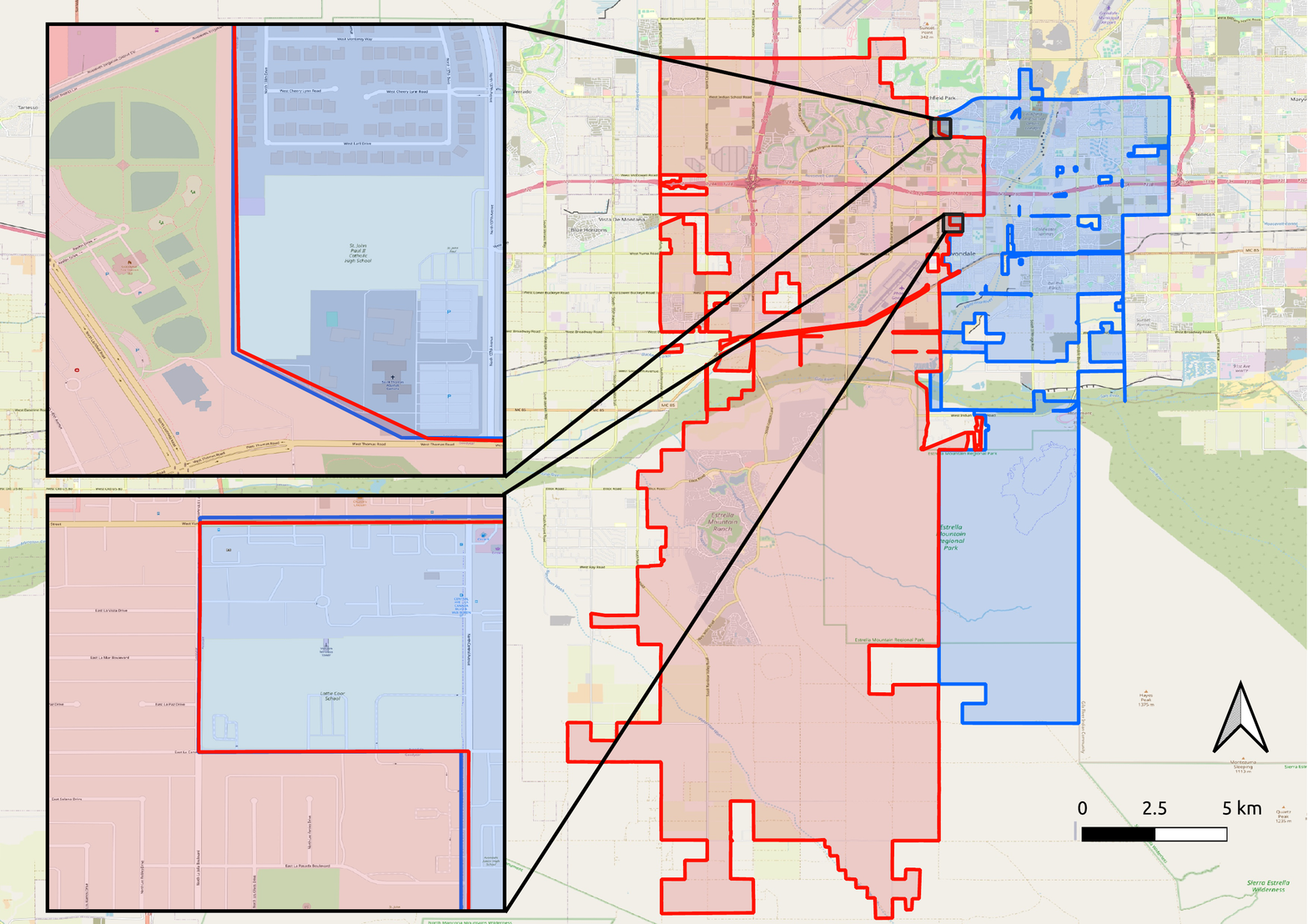}
        \caption[]%
		{Polygon Touch Error
		}
        \label{fig:touch}
    \end{subfigure}
        \caption{Three examples to show the geometry uncertainty problem in OpenStreetMap: 
        (a) Lynchburg, Tennessee (the red polygon) is a consolidated city-county whose boundaries are identical to Moore County, Tennessee (the blue polygon). However, the answer to Question \textit{Is Lynchburg, Tennessee equivalent to Moore County, Tennessee} is No, if we compute the spatial \revise{relation} between these two polygon geometries based on GeoSPARQL function \textit{geof:sfEquals}.
        (b) Powellton, West Virginia (the red polygon) is a census-designated place inside of Fayette County, West Virginia (the blue polygon). However, for Question \textit{is Powellton, West Virginia inside of Fayette County, West Virginia}, the answer is No if we use GeoSPARQL function \textit{geof:sfWithin} to compute the spatial relation between their polygon geometries.
        (c) Avondale, Arizona (the blue polygon) is a nearby city of Goodyear, Arizona (the red polygon). However, As for Question \textit{Does Avondale, Arizona touch Goodyear, Arizona} or Question {Is Avondale, Arizona externally connected to Goodyear, Arizona}, if we compute the answer based on GeoSPARQL function \textit{geof:sfTouches}, their answers are both No because their OpenStreetMap polygons intersect with each other.
        }
    \label{fig:pgonfunc--err}
\end{figure*}

Figure \ref{fig:pgonfunc--err} uses three examples from OpenStreetMap to show the problem of geometric uncertainty. Each of these three examples consists of a pair of geographic entities who are represented by a red polygon and a blue polygon. By using the Region Connection Calculus 8 (RCC8) \citep{cohn1997qualitative}, the expected spatial relations between these three pairs are \textit{equal} (OE), \textit{tangential proper part} (TPP), and \textit{externally connected} (EC) respectively. However, because of the geometric uncertainty, if we compute their spatial relations based on their polygonal geometries, in all these three examples, their spatial relations become \textit{partially overlapping} (PO). 
As shown in those zoom-in windows in Figure \ref{fig:ispartof} and \ref{fig:touch}, these unwanted small polygons which break the topological relations between regions are also called ``sliver polygon''\footnote{\url{https://en.wikipedia.org/wiki/Sliver_polygon}}. 
For example, in Figure \ref{fig:ispartof}, Powellton, West Virginia (the red polygon) should be a subdivision of Fayette County, West Virginia (the blue polygon). However, because of the small sliver polygon shown in the enlarged window, their relations become \textit{partially overlapping} (PO) if we strictly compute the spatial relation based on their geometries and without pre-processing, e.g., by using GeoSPARQL spatial relation functions~\citep{battle2012enabling}.

\citet{regalia2019computing} also recognized the effect of geometry uncertainty on the spatial relationship computation. To overcome this problem, \citet{regalia2019computing} proposed to precompute metrically-refined topological relations \citep{egenhofer2009topological} between geographic entities and materialize them as triples in a geographic knowledge graph. So a GeoQA system only needs to do triple lookup for question answering instead of computing topological relations on-the-fly. However, except for the problem of a substantial larger triple set, how to decide thresholds for metrically-refined topological relation computation is still a big question since these thresholds vary according to the geographic feature types under consideration and the map scale of these geometries. 

% A more proper way is to design an efficient neural-network-based ``fuzzy spatial operator'' which is robust to the geometric uncertainty problem.
% This ``fuzzy spatial operator'' takes these complex polygon geometries as input and output their spatial relations. Based on back-propagation, this operator automatically learns the concept of thresholds implicitly based on the training labels and we does not need to specify thresholds explicitly. This might be an interesting research direction.

\subsubsection{Vagueness of Geographic Concepts and Entities}  \label{sec:geo-vague}
However, even if we can fix the problem of geometric uncertainty, a GeoQA system can still fail to answer many geographic questions because of the inherent vagueness of many geographic concepts such as forest, lake, desert, swamp \citep{bennett2002forest,kuhn2003semantic}, or even coastline. For instance, aside from the geometric uncertainty when digitizing the coastline of Great Britain, the concept ``coastline'' is conceptually vague. The exact coastline of Great Britain varies according to the time of the day and the season when we measure it. The spatial extent of Amazon forest really depends on the definition of ``forest'' and can be potentially controversial. \citet{bennett2002forest} has listed 12 main
aspects of vagueness associated with the term ``forest'' such as \textit{How dense must the vegetation be} and \textit{How large an area must a forest occupy}. Given \revise{the} vagueness of geographic concepts, it is \revise{particularly} challenging to pick a correct spatial representation for a geographic entity associated with these concepts. So answering geographic questions that \revise{involve} these concepts is prone to errors, such as \textit{How many lakes there are in Michigan}, \textit{What is the total area of Amazon forest}, \textit{How far it is from Rocky Mountain to Denver}, and so on.

Interestingly, the vagueness of a geographic entity can not only come from its vaguely defined geographic feature types/concepts, but also come from its own definition such as vague cognitive regions \citep{montello2014vague}. Good examples are Downtown Santa Barbara \citep{montello2003s} and Northern California \citep{montello2014vague,gao2017data}. It is hard to represent their spatial footprints as polygons with crisp boundaries. Instead, they are usually represented by fuzzy boundaries or membership scores. Answering geographic questions involving these kind of entities is also challenging, i.e., \textit{Is San Luis Obispo part of Southern California?}

.

\section{Uniqueness of Geographic Questions and GeoQA} \label{sec:unique}

%Although the discussions in Section \ref{sec:geoqa-problem} are based on three pairs of geographic questions shown in Figure \ref{fig:challenge-geoqa}, they already cover a wide range of problems and challenges we meet when designing a GeoQA system. 
Based on the above discussions, the key challenges of GeoQA are summarized as follows. % (some of which are shared with other QA systems). 
Some general challenges are shared with other QA systems:
\begin{enumerate}
    \item \textbf{Linguistic variability}: the same question can be expressed in different ways. Paraphrase, hyponym, and synonymy cause a large linguistic variability of (geographic) questions \citep{berant2013semantic}.
    \item \textbf{Program variability}: there are many possible \textit{programs}\footnote{In semantic parsing and structured data source QA research~\citep{pasupat2015compositional,liang2017neural}, programs indicate queries such as SPARQL queries, SQL queries, and $\lambda$-calculus \citep{yih2015semantic} which are \textit{translated} from natural language questions and can be executed on the underlining knowledge base to retrieve the answer.} \citep{liang2017neural} to answer a given (geographic) questions and each of them are correct. This increases the search space and makes a QA model difficult to train.
    \item \textbf{Question complexity}: there are various types of geographic questions \citep{punjani2018template,hamzei2019place}. Different question types require different data sources and QA techniques to represent the answer. In the first step, it is better to narrow down the scope of the QA systems, i.e., the types of questions the QA system can handle.
    \item \revise{\textbf{Data source diversity}: there are various data sources which can be used as knowledge bases for QA such as knowledge graphs, semi-structured tables, text corpus. Sometimes it is necessary to answer questions based on multiple data sources. It becomes more demanding in the GeoQA context since most geographic questions have to be answered based on a combination of multiple data sources such as raster data, vector data, text corpus, geographic knowledge graphs, and so on. Hence, developing QA systems based on multiple data sources is particularly challenging.}
\end{enumerate}

% \begin{enumerate}
%     \item There is no large-scale and diverse geographic question answering benchmark dataset to train deep-learning-based QA models. Geoquery \citep{zelle1996learning}, GeoQuestion201 corpus \citep{punjani2018template} are rather small data sets which contains 500 and 201 geographic questions respectively. MS MARCO V2.1 data set \citep{hamzei2019place} has limited question types.
%     \item Vague concepts, vague entities, and vague spatial relations.
%     \item  There is no a well-defined set of spatial operators to execute in order to answer geographic questions.
% \end{enumerate}

There are unique challenges which are specific for geographic question answering. Based on Section \ref{sec:geoqa-problem} and \citet{mai2019relaxing}, these unique challenges can be summarized as follows:

\begin{enumerate}
    \item Answering geographic questions relies on \textbf{appropriate spatial information} such as geometries (e.g., points, polylines, and polygons). Inappropriate selection of spatial footprints will lead to wrong answers as shown in Figure \ref{fig:qa1-1} and \ref{fig:qa1-2}.
    
    \item A GeoQA system should be robust in handling \textbf{the vagueness and uncertainty of geographic information}. For example, a lake can have different definitions and different polygonal representations at different map scales. These uncertainties and vagueness might change the spatial relations between these polygon geometries as shown in Figure \ref{fig:pgonfunc--err} and discussed in Section \ref{sec:geo-uncertain-vague}. A GeoQA system should be able to handle this. % type of uncertainty.
    
    \item Answers to many geographic questions are best derived from \textbf{a sequence of spatial operations} such as proximity (Figure \ref{fig:qa1-1}, Figure \ref{fig:qa1-2}), topological and cardinal direction (Figure \ref{fig:qa2-1}, Figure \ref{fig:qa2-2}), and routing computation rather than being directly extracted from a piece of unstructured text~\citep{asai2020learning} or retrieved from Knowledge Graphs (KG)~\citep{berant2013semantic}, which are the normal procedures in current QA systems.
    
    \item Compared with the general QA, answering geographic questions requires \textbf{a substantially larger set of programs/operators}, especially a large set of spatial operators. This increases the program search space exponentially. For example, PostGIS has 21 spatial relationship functions (e.g., ST\_Within), 27 measurement functions (e.g., ST\_Azimuth), and 25 geometry processing functions (e.g., ST\_Buffer) \footnote{\url{https://postgis.net/docs/reference.html}}. In contrast, in the general QA research, the current semantic parser \citep{yih2016value,liang2017neural} or reading comprehension QA \citep{chen2020neural} usually only utilize a small set of operators to make the whole model trainable.
    For instance, 
    %in order to answer natural language questions from the WebQuestionsSP dataset \citep{yih2016value} based on the Freebase knowledge graph, 
    Neural Symbolic Machine (NSM) \citep{liang2017neural}, as a neural sequence-to-sequence semantic parser, automatically translate a question into a program that can be executed on the KG and retrieve answers with the support of a Lisp interpreter. This Lisp interpreter only supports 4 operators - \textit{Hop}, \textit{ArgMax}, \textit{ArgMin}, and \textit{Filter}. Neural Symbolic Reader (NeRd) \citep{chen2020neural}, as a scalable reading comprehension QA, only supports 11 different operators. The total number of possible programs that can be generated grows exponentially with respect to the number of operators we consider. So the large number of spatial operators makes this program generation task \revise{extremely} complex.
    
    \item Geographic question answering can be \textbf{subjective and context dependent}, i.e., depending on when and where this question is asked, who ask it, and what this question is asked about. Some examples are \textit{Is California (the territory) part of the United States} (time-dependent), \textit{which country contains the largest proportion of the Kashmir region} (location-dependent and subject-dependent). The answer to the first question can be USA or Mexico depending on the temporal scope of this question. The answer to the second question can be India or Pakistan depending on when, where, and who you ask this question \citep{janowicz2015data}. 
    
    \item Geographic questions can be \textbf{vague in terms of the involved spatial relations and geographic concepts}. For instance, the answer to Question \textit{In what direction is France located to Italy} can be either east or southeast depending on the definition of cardinal directions between polygons. \revise{Moreover}, for Question \textit{\revise{What is the total area of forest in Brazil}}, the answer depends on the definition of forest \citep{kuhn2003semantic}.
\end{enumerate}

\section{Existing Work on GeoQA} \label{sec:related}
%Based on the previous discussion, it is clear that geographic question answering is an interesting yet challenging task to tackle. 
Although QA has been a long-standing research topic, geographic question answering (GeoQA) remains less studied. %nearly untouched. 
In this section, we discuss some important existing work on GeoQA. Based on the types of geographic questions they focus on, we classify existing GeoQA research into four types: factoid, geo-analytical, scenario-based, and visual. 

\subsection{Factoid Geographic Question Answering} \label{sec:related-fact}
Factoid GeoQA focuses on answering questions based on geographic facts. To the best of our knowledge, \citet{zelle1996learning} presented the first GeoQA system, which uses CHILL parser to answer natural language geographic questions based on the \textit{Geoquery} query language. They defined 20 relations such as \textit{capital}, \textit{area}, \textit{next\_to}, \textit{traverse}, and so on, which indicate different types of geographic questions that Geoquery supports. Although some relations are spatial such as \textit{next\_to} and \textit{traverse}, all relations have been materialized as 800 Prolog facts. Then the QA system only needs to perform a question-query translation and an answer lookup. Namely, no on-the-fly spatial computation is required. Although this work focused on answering geographic questions, a standard QA pipeline was adopted and the uniqueness of geographic questions was not considered.

\citet{chen2013synergistic} proposed a geographic question answering framework to answer five types of geographic questions based on the spatial operators supported by PostGIS. An input geographic question first goes through a linguistic analysis so as to be classified into one of the predefined query templates. Then the spatial SQL query template is filled by using the parsed data such as spatial operators (e.g., ST\_Within, ST\_Buffer), place name, quantity constraints, and so on. Subsequently, the answer is retrieved by executing this query on the underlining PostGIS database. This GeoQA framework can support five simple geographic question types: 1) location questions, e.g., \textit{where is Columbus}; 2)  direction \& distance questions, e.g., \textit{where is Columbus perspective to Cleveland}; 3) distance questions, \textit{how far is it from Columbus to Cleveland}; 4) nearest questions, e.g., \textit{which city is the nearest to Columbus}; 5) buffer questions, e.g., \textit{which cities are within 5 miles from Columbus}. We can see that except for the first type of questions, the rests require spatial operators. Compared with \citet{zelle1996learning} who materialized all spatial relations as facts beforehand, this system is able to utilize spatial operators to answer geographic questions on-the-fly. However, it simply utilizes points to represent geographic entities and thus inherits the limitation we have discussed in Section \ref{sec:geoqa-problem}. The limited number of question types and the small size of the underlying database restrict the number of geographic questions it can handle.

\citet{punjani2018template} proposed a template-based GeoQA system as \citet{chen2013synergistic} did. Instead of relying on a PostGIS database, this GeoQA system is based on a GeoSPARQL-enabled geographic knowledge graph created from DBpedia, GADM database of global administrative areas, and OpenStreetMap. This GeoQA system mainly focuses on seven types of factoid geographic questions which can be answered based on several handcrafted GeoSPARQL query templates. These question types include various numbers of geographic entities, concepts, or spatial relations. First, geographic entities, concepts, and spatial relations are extracted from a natural language geographic question asked by users. Then this question is mapped to one of the query templates. The generated GeoSPARQL query is then executed on the underlining KG to obtain answers. This GeoQA system is able to handle different spatial relations such as topological relations and cardinal direction relations by using the polygon geometries of each geographic entity. However, the deterministic spatial operations supported by GeoSPARQL suffer from uncertainty of the polygon geometries we have discussed in Section \ref{sec:geo-uncertain}.

As a prerequisite of GeoQA, \citet{hamzei2019place} carried out a data-driven place-based question analysis using a large-scale QA dataset generated from Microsoft Bing - MS MARCO V2.1. They used linguistic analysis to translate questions and answers into their semantic encodings based on six primary elements: place names, place types, activities (e.g., buy), situations (e.g., live), qualitative spatial relationships, and qualities. Then they used a string similarity measure (Jaro similarity) as well as k-means to cluster the encoded questions and answers into different clusters. Experimental results 
\revise{showed} that place-based questions can be clustered into three types: 1) non-spatial questions - questions not aiming at localization of places (e.g., \textit{In which county is Grand Forks, North Dakota located}); 2) spatial questions - questions about locations of place (e.g., \textit{where is Barton County, Kansas}); 3) non-geographical and ambiguous questions (e.g., \textit{where are ores located}). The proposed semantic encoding approach benefits our understanding of the intent of geographic questions. However, this classification is rather coarse. The non-spatial question type still contains various types of factoid geographic question.
Moreover, this classification is still based on the syntactic structures of questions rather than their semantic interpretations.  The geographic question types discussed in \citet{hamzei2019place} are only factoid questions. In contrast, we provide a classification of geographic questions in Section \ref{sec:geoqaclass} based on their semantic interpretations which cover a wider range of question types.

Based on the above discussion, we can see that although there is some research on factoid GeoQA, most existing GeoQA models \citep{zelle1996learning,chen2013synergistic,chen2014parameterized,punjani2018template} are template-based and can only handle limited types of geographic questions. Commonly, they \revise{adopted} a two-step strategy to answer geographic questions -- a question classification step and an answering step. A natural language question is first classified into one predefined query template, which then is used in QA system to seek the answer.   
This indicates that, these models are not directly trained on the labeled data, namely question-answer pairs. Instead, they are usually trained on the intermediate question type labels which does not guarantee for a correct final answer while this error cannot propagate back to the whole QA framework.
%Lots of efforts have been invested to design query templates and to build question classification models . However, these question classification models are not directly trained on the labeled data, namely question-answer pairs. Instead, they are usually trained on the intermediate question type labels which does not guarantee for a correct final answer while this error cannot propagate back to the whole QA framework.
%it is possible that the generated query cannot retrieve the correct answer and this error cannot propagate back to the whole QA framework.
Therefore, existing GeoQA models can hardly be trained in an end-to-end manner as many reading comprehension QA models do \citep{liang2017neural,asai2020learning} and cannot be easily generalized to other datasets as well. In short, there is still a lack of efficient large-scale end-to-end GeoQA systems which can handle various types of geographic questions. 

\subsection{Geo-analytical Question Answering} \label{sec:related-geoanalytic}
Compared with the above GeoQA work that mainly focus on answering factoid geographic questions, geo-analytical question answering proposed by \citet{scheider2020geo} \revise{went} beyond simple geographic facts but focuses more on questions with complex spatial analytical intents \citep{xu2020extracting}. A simple factoid geographic question such as Question A1 can be answered by executing one or two spatial operations on the respective spatial footprints of geographic entities. In contrast, geo-analytical questions usually require generating a GIS analytic workflows.
Example questions include \textit{how much green space will Tom see while running through Amsterdam} (Question M) \citet{scheider2020geo} and \textit{what is the best site for a new landfill in the Netherlands} (Question N) \citep{xu2020extracting}.

The aim of geo-analytical question answering also shifts from retrieving simple answers to formulating the answer through analytical workflows which might be generated on-the-fly or retrieved from a GIS workflow corpus shared by other GIS users \citep{scheider2019finding}.

Despite the interesting nature of geo-analytical QA, several challenges need to be solved in order to develop a full-functional geo-analytical QA system. Firstly, in contrast to all current QA systems which are built on \textit{predefined knowledge bases} (e.g., knowledge graphs, text corpus, and semi-structured tables), \revise{geo-analytical question answering does not have well-defined knowledge bases. 
%it is unclear how to design a knowledge base that can be used to support geo-analytical QA. 
Different geo-analytical questions might require different kinds of knowledge bases. %And for some geo-analytical questions, it is unclear what kind of knowledge base can be used to answer them. 
\citet{scheider2020geo} turned to treat a portal of different GIS datasets as the knowledge base of geo-analytical QA. However, all current Geoportals such as ArcGIS Online \citep{hu2015metadata,mai2020semantically} and NASA Physical Oceanography Distributed Active Archive Center (PO.DAAC) \citep{jiang2018towards} only support search functionality over different datasets on the metadata level and cannot be directly used for geo-analytical QA which requires a deep assessment of the analytic potential of a GIS dataset for a given question.
%For example, as for Question M above, we can use a vector layer of trees in Amsterdam or a raster layer of the green space in Amsterdam to compute the answer. However, none of these datasets is considered in a normal QA system and including vector and raster data as another knowledge base for GeoQA is challenging.
} 
Secondly, geo-analytical questions are mostly vaguely defined and can be answered based on different combinations of data sets and GIS tools (spatial operators). 
For example, as shown in \citet{scheider2020geo}, to answer Question M, one option is to use a vector map of urban trees in Amsterdam overlaid on Tom's running trajectory, based on which the number of trees within the buffer of the trajectory can be computed to answer the question. Another option is to use a raster map of green space in Amsterdam and computing the answer based on kernel density estimation and map algebra. Different data set options make it difficult to design a knowledge base for geo-analytical QA. Different possible solutions lead to a growing solution space and therefore make it harder to construct a fully automatic QA pipeline. 
It is these difficulties that make geo-analytical QA challenging and worth investigating at the same time.

\subsection{Scenario-based Geographic Question Answering}  \label{sec:related-scenario}
In scenario-based GeoQA (GeoSQA), a question is always associated with a scenario described by a map or a paragraph.
\citet{huang2019geosqa} presented a GeoSQA dataset which consists of 1,981 scenarios and 4,110 multiple choice questions in geography domain. These scenarios and multiple choice questions are collected from Gaokao, China's version of SAT, and mock tests at high school level. So all scenario-based geographic questions are textbook-like questions. An example scenario can be a map showing the urban planning of a city as well as its textual description. The associated question asks for the possible usage of a location presented on the map. Answering this kind of questions requires some commonsense knowledge in geography as well as a deep understanding of the scenario. \citet{huang2019geosqa} showed that the state-of-the-art reading comprehension and textual entailment models perform no better than random guess on this task which illustrates the challenges of this kind of GeoQA.

In contrast to the textbook-like scenario-based QA, \citet{contractor2020joint} presented a tourism oriented scenario QA task and a GeoQA pipeline. The target QA dataset - Tourism Questions \citep{contractor2019large} consists of over 47,000 real-world tourism questions that seek for Points-of-Interest (POI) recommendations together with a universe of nearly 200,000 candidate POIs. These questions are long paragraphs which describe a tourism scenario asking for POI recommendation. An example question is \textit{I am outside of Universal Studio, Los Angels, please recommend good Chinese restaurants nearby}\footnote{Since the original question example is very long. We formulate a rather short and simple tourism scenario question here.}. The answer to these questions are usually a ranked list of POIs. To tackle this task, \citet{contractor2020joint} proposed a spatio-textual reasoning network which jointly considers the spatial proximity between candidate POIs and the target POIs in the question as well as the semantic similarity between questions and the reviews of candidate POIs. The distances between candidate POIs and the target POIs mentioned in the question are explicitly encoded by a geo-spatial reasoner module which produces the spatial relevant scores between questions and candidate POIs. The semantic relevant scores are computed by a textual reasoning sub-network. These two scores are then combined to produce the final relevant scores between questions and each candidate POI. The approach indeed shows a great potential of spatial reasoning in GeoQA. However, since distances need to be computed for each pair of candidate POIs and target POIs in the questions, the presented spatio-textual reasoning network is not suitable for open-domain QA where we can have a richer pool of candidate POIs to search from.

\subsection{Visual Geographic Question Answering} \label{sec:related-visual}
Visual question answering \citep{antol2015vqa} is another rapidly developing QA research direction in which each question is paired with an image as the context. An example question that could be asked about an image showing a child is \textit{where is the child sitting}. \citet{lobry2020rsvqa} adopted this idea and proposed the task of visual question answering for remote sensing data (RSVQA) in which a remote sensing image is paired with a question asking about the content of this image. Example questions include \textit{how many buildings are there}, and \textit{what is the area covered by small buildings}. To answer this kind of questions, \citet{lobry2020rsvqa} utilized a Convolutional Neural Networks (CNN) as the image encoder and a Recurrent Neural Network (RNN) as the question encoder. The encoder outputs are concatenated and fed to a fully connected layer which is followed by an answer classification layer. Although this work mainly focuses on capturing computer vision features, spatial knowledge is minimally utilized in the RSVQA model design. Consequently, the presented RSVQA model shows little difference compared to normal VQA models. How to incorporate spatial thinking into the RSVQA model design to develop spatially-explicit \citep{janowicz2020geoai} QA models is a promising future research direction.

\section{The Classification of Geographic Questions} \label{sec:geoqaclass}

Section \ref{sec:related} discussed key work on GeoQA which focus on certain types of geographic questions. Some of them \citep{chen2013synergistic,punjani2018template,hamzei2019place,xu2020extracting} provided a classification of geographic questions within the scope of question types they can handle.
In this section, we provide a general classification of geographic questions which attempts to cover all aspects of GeoQA. We hope this classification can comprehensively reveal the landscape of GeoQA and serve as a guideline for future GeoQA-related research.

In fact, \citet{mishra2016survey} provided a survey for question answering systems and classified QA systems based on multiple criteria including application domains, question types, types of analysis on questions, types of data sources, retrieval methods, and answer types. According to \citet{mishra2016survey}, questions can be classified as factoid type questions [what, when, which, who, how], list type questions, hypothetical type questions, causal questions [how and why], and confirmation questions. Although the classification covers most of the questions asked in a normal QA system, it does not consider many important types that we often see in geographic questions such as questions about spatial relations, routing questions, prediction-based questions, and so on.

Following classification by \citet{mishra2016survey}, we classify geographic questions into the following categories: %based on their question intents, the information associated with questions, the answer types, and the way how GeoQA systems might generate answers.
\begin{enumerate}
    \item \textbf{Factoid geographic questions}: geographic questions that can be answered based on the factoid geographic knowledge, e.g., \textit{which state is Houston located in}.
    \item \textbf{Prediction-based geographic questions}: geographic questions should be answered based on the prediction of facts, e.g., \textit{what will be the average temperature in Las Vegas next Monday}.
    \item \textbf{Opinion geographic questions}: geographic questions which require subjective information or opinions about some geographic facts, e.g., \textit{what is the best trail in the Grand Canyon National Park}.
    \item \textbf{Hypothetical geographic questions}: geographic questions that ask for information related to any hypothetical events, e.g., \textit{what would California look like if the United States had not acquired it in 1848}.
    \item \textbf{\revise{Causal} geographic questions}: geographic questions which require explanations about geographic facts, e.g., \textit{why and how did Los Angeles become famous for its film industry}.
    \item \textbf{Geo-analytical questions}: geographic questions which require complicated geoprocessing workflows to answer, e.g., \textit{where is the best location for my new house in San Diego with a quiet neighborhood, lower crime rate, good accessibility to grocery stores and beach}.
    \item \textbf{Scenario-based geographic questions}: geographic questions that are associated with a scenario described by textual description or a map. An example question is \textit{we just arrived at London and currently stay at a hotel close to London King's Cross train station. Can you recommend a good Italian restaurant nearby which serves vegan pizza?}
    \item \textbf{Visual geographic questions}: geographic questions paired with remote sensing images or maps whose contents are the focus of these questions. 
\end{enumerate}

In the following, we will discuss each question type in detail.

\subsection{Factoid Geographic Questions} \label{sec:factiodqa}
In contrast to the factoid type questions defined by \citet{mishra2016survey} that require answers in a single short phrase or sentence and whose expected answer types are named entities, we define factoid geographic questions in a broader sense in terms of the answer types. Any questions that can be answered based on the real-world factoid geographic knowledge can be treated as factoid geographic questions. The factoid type questions and list type questions\footnote{list type questions are questions whose answer are a list of entities. This question type is still based on factoid knowledge.} \citep{mishra2016survey} are included in this question type if they are geographic questions. 

Factoid geographic questions are the most typical question type that existing GeoQA systems focus on. We further classify this type into the following sub-types:

\begin{enumerate}
    \item \textbf{Single geographic entity attribute questions}: This type refers to questions about attributes of one single geographic entity such as its geographic coordinates, population, elevation, area, temperature, and so on. 
    %This kind of questions can be answered by the triples with object type properties in a geographic knowledge graph or descriptions or reviews of a place.
    This question type does not require any spatial operations and thus can be answered via a datatype property\footnote{\url{https://www.w3.org/TR/owl-ref/\#DatatypeProperty-def}} triple fetched from a GeoKG or extracted from a description of a place. Examples include \textit{where is London}, \textit{what is the total population of Phoenix, Arizona}, and \textit{what is the annual precipitation in Seattle, Washington}.
    
    \item \textbf{Spatial relationship questions}: These are questions that involve spatial relations such as spatial proximity, topological relations, cardinal directions, ternary projective relation, and n-ary spatial relations between/among (two or more) geographic entities. Examples of this type include: \textit{how far is it from New York to Washington D.C.} (spatial proximity), \textit{how much does it cost to take a Uber from Stanford University to Pier 33} (time dependent spatial proximity), \textit{does King Canyon National Park touch Inyo County, California} (topological relations), \textit{What is the cardinal direction between Los Angles and San Diego} (cardinal directions), \textit{which country sits between China and Russian} (ternary projective relation), and \textit{which countries surround Switzerland} (n-ary spatial relation).
    
    \item \textbf{Spatial/non-spatial qualifier questions}: This refers to those questions that are asked about one or a set of geographic entities which satisfy one or several spatial (e.g., in City A) or non-spatial (e.g., highest elevation) qualifiers. Examples include: \textit{What is the largest city in United States in terms of population}? \textit{Which province in China has the highest average elevation}? \textit{Which coastal cities are within 20 miles from Seattle}? \textit{Which churches are near a castle in Scotland}, and \textit{Which city in France has the largest COVID-19 case count}.
    
    \item \textbf{Routing questions}: This type of questions is frequently asked in navigation guidance services and mainly asks about the routing between places. The answer is, therefore, a route displayed on the map or a voice/text-based step-by-step instruction. One example is: \textit{how to get from Hollywood to LAX airport?}
    
\end{enumerate}

These sub-types comprehensively cover geographic question types that have been discussed in \citet{zelle1996learning,chen2013synergistic,chen2014parameterized,punjani2018template,hamzei2019place,xu2020extracting}. %Although factoid geographic questions are relatively simple compared with other question types, there are still a lot of challenges to be solved as we have discussed in Section \ref{sec:geoqa-problem}.

\subsection{Prediction-based Geographic Questions} \label{sec:predictqa}
Factoid geographic questions ask about historical or present geographic knowledge, while prediction-based geographic questions ask about the future. Hence answers should be generated based on predictions of real-world geographic facts such as population, temperature, future events, and so on. 

In some cases, the predictions have been precomputed and stored in a knowledge base. %such as future temperature, air pollution, predicted COVID-19 affected cases.
Then the QA process of prediction-based geographic questions can be done in exactly the same way as that of factoid geographic questions. 
%So we can do a similar question classification for prediction-based geographic questions. Since the spatial relations and routing will not be changed in a fairly long time, we leave out some question sub-types and 
We classify prediction-based geographic questions as follow:

\begin{enumerate}
    \item \textbf{Single geographic entity attribute prediction question}: Questions about the prediction of attributes of one single geographic entity such as population, air quality, temperature, and so on. e.g., \textit{what will be the air quality like in Los Angels in the following two weeks}, \textit{where this iceberg will be in two months after its recently separation from the Antarctic glacier}.
    
    \item \textbf{Spatial/non-spatial qualifier prediction questions}: Questions asked about one or a set of geographic entities which satisfy one or several spatial or non-spatial qualifiers in the future.
    \begin{enumerate}
        \item \textbf{Prediction-based non-spatial qualifier questions}: These questions have non-spatial qualifiers which are based on the predictions of the attributes of geographic entities. Examples are \textit{which country in the world will have the largest population in 10 years}, \textit{which state in the US will have the largest total COVID-19 case count once this current pandemic ends}, \textit{which university in Australia will have the largest proportion of international students in 5 years}.
        
        \item \textbf{Prediction-based spatial qualifier questions}: These prediction questions have spatial qualifiers for geographic entities whose locations may or may not change, %have spatial qualifiers which are based on predictions of the change of locations of geographic entities, 
        e.g., \textit{which nearby house will have the largest increase in its price after the construction of this subway station that will be finished in two years}.
    \end{enumerate}
\end{enumerate}

If the predictions are not available beforehand, the GeoQA system should be able to understand the question intent and generate a program to compute the answer which might involve some prediction functions. As far as we know, there are no QA systems available to date that address this type of GeoQA.

\subsection{Opinion Geographic Questions} \label{sec:opinion}
Opinion geographic questions involve personal opinions with subjective terms such as the \textit{best} hotels, the \textit{most beautiful} city, the \textit{most atmospheric} restaurant, and so on. These subjective terms can be interpreted in different ways by different people which complicates the question answering process. For example, as for Question \textit{what is the largest city in Texas}, ``the largest city'' can be interpreted as the city with the largest population, or with the largest area. Some subjective terms can be approximated based on existing quantitative measures. For example, as for Question \textit{what is the most popular restaurant in San Jose, California}, we can use the Yelp rating as a proxy to measure the popularity of a restaurant. In this case, the QA can be done in the same way as that of the factoid geographic questions. Nevertheless, opinion detection itself which classifies text as subjective or objective is still a research problem \citep{khan2014mining}.

\subsection{Hypothetical  Geographic  Questions} \label{sec:hypoqa}
Similar to the definition provided by \citet{mishra2016survey}, hypothetical  geographic  questions ask for information related to any hypothetical event or condition. The question is usually formulated as ``what would happen if...''. Example questions are \textit{what would California look like if the United States had not acquired it in 1848}, \textit{which nearby cities would have been flooded if the dike at Huayuankou, Henan would have been breached again}\footnote{\url{https://en.wikipedia.org/wiki/Huayuankou,_Henan}}.

At first glance, hypothetical geographic questions might look similar to prediction-based geographic questions. However, they are different question types. The former asks for a hypothetical situation and the answers are usually derived from an educated guess based on commonsense. In contrast, the later asks for a scientific prediction based on the observation data.
%The former asks for a hypothetical event which might not happen in the future while the later asks for predictions of the fact which will happen in the future.

Since there are no 100\% correct answer for these questions, the QA reliability is low and the QA technique adopted by factoid question answering will not work. Some expert knowledge and commonsense knowledge may need to be involved during the QA process. This question type might be one of the most difficult one to handle and need to be investigated further.

\subsection{\revise{Causal} Geographic Questions} \label{sec:causalqa}
\revise{Causal} geographic questions ask for explanations about geographic facts. Example questions are \textit{why are there a lot of places along the west coast of the Atlantic Ocean named after Alexander von Humboldt} and \textit{why are there a lot of places in South America named San Jose}.

The answer to a \revise{causal} geographic question is usually a passage about the geographic facts under discussion. So we can adopt some text-corpus-based extractive question answering techniques \citep{chen2017reading,karpukhin2020dense} to ``approximately'' answer this kind of questions. 
However, almost all the current deep neural network based extractive QA models \citep{chen2017reading} \footnote{Given a question, an extractive QA model search for the possible paragraphs which might contain the answer. And then it reads these paragraph sand \textit{extract} text spans from them as the answers.} can only do fact lookup from text corpus while \revise{causal} geographic questions require a deep understanding of the causality relationship in the questions and reasoning on commonsense knowledge. So simply applying extractive QA models on \revise{causal} questions will lead to much lower performance. 
%without a clear understanding of the question intent by the deep neural network, the recall might be lower than the normal extractive QA.

\subsection{Geo-analytical  Questions} \label{sec:geoanlyticqa}
Section \ref{sec:related} provides a detailed description of geo-analytical QA and discusses about the challenges we might meet when developing a geo-analytical QA system - uncertain choices of knowledge bases and exploded solution space.

The reasons why we separate geo-analytical questions from other types of questions are two-fold: 1) unlike other types of questions that aim at generating compact answers, geo-analytical question answering focuses more on generating or retrieving the geoprcessing workflows \citep{scheider2020geo} that can be used to obtain answers; 2) in contrast to other question types that have relatively limited answer types, the answer types of geo-analytical questions are very diverse. Example answer types include raster maps, geometries, numerical values, geographic entities, text, and so on.

Despite its difficulty, geo-analytical QA actually points out an exciting future direction of GIS technology which can automate the spatial analysis process without any human intervention. 
%Despite its difficulty, 
So we still advocate this idea and expect a major advancement along this research direction in the early future.

\subsection{Scenario-based Geographic Questions} \label{sec:scenarioqa}
As we discussed in Section \ref{sec:related-scenario}, a scenario-based geographic question is usually associated with a scenario depicted by either a map or a textual description. Classical scenarios used in GeoQA include the textbook-like scenario such as the GeoSQA dataset \citep{huang2019geosqa} and the tourism scenario such as Tourism dataset \citep{contractor2019large,contractor2020joint}. 
As for Tourism datasets \citep{contractor2019large}, only simple spatial reasoning, e.g., distance between candidate POIs and POIs mentioned in the scenario, is required. However, as for the GeoSQA dataset \citep{huang2019geosqa}, different textbook scenarios require different spatial reasoning such as cardinal directions, proximity, and topological reasoning. 
Moreover, commonsense knowledge is required to correctly answer this type of questions. 
Therefore, designing a spatial-aware QA model for GeoSQA is challenging. 

So for scenario-based geographic questions, the design of GeoQA model varies from case to case and depends on the nature of the questions and what scenario the questions are based on.

\subsection{Visual Geographic Questions} \label{sec:visualqa}
Visual geographic questions are different from other question types because each question is paired with a remote sensing image \citep{lobry2020rsvqa} or a map. These images or maps can be seen as the restricted knowledge base for corresponding questions. The map can be a historic map or a narrative map. They can also be obtained from some fictions, such as Marauder's Map from Harry Potter, 
\revise{\textit{Atlas of the European novel, 1800-1900} \citep{moretti1998atlas}, and \textit{A Literary Atlas of Europe}\footnote{\url{http://www.literaturatlas.eu/en/}}.}
%Moreover, a map can also be a vector layer such as a polygon shapefile in which the corresponding QA process should be similar to semi-structured QA \citep{berant2013semantic} but adding more spatial operations. 
%\revise{In narrative cartography, there are some research studying the geography of the fictions such as \textit{Atlas of the European novel, 1800-1900} \citep{moretti1998atlas} and the research project \textit{A Literary Atlas of Europe}\footnote{\url{http://www.literaturatlas.eu/en/}}. 
\revise{However, to the best of our knowledge, these narrative maps have not been used for the GeoQA purpose and there is no visual GeoQA work focusing on fictional maps.}

Promising research questions for visual geographic questions answering include issues such as what makes visual GeoQA different from normal visual QA? What are the benefits to incorporate spatial knowledge into Visual GeoQA models?
%Although we have discussed the uniqueness of GeoQA in Section \ref{sec:unique}, they are all applied to other types of geographic questions. 
One possible direction lies in the difference between the spatial relations used in general VQA and geographic VQA. The spatial relations studied in the current VQA \citep{ramalho2018encoding} are like \textit{on the left of}, \textit{in front of}, and \textit{on top of} which is very different from the spatial relations we would have among geographic entities, e.g. cardinal direction, topological relations. Whether this difference leads to some difference in the GeoQA model design needs to be investigate further.

\subsection{Discussion about the Question Classification} \label{sec:classdiscuss}

\revise{The proposed question classification is an integration and extension of multiple existing question classification work \citep{mishra2016survey,punjani2018template,hamzei2019place}. 
In fact, these question types are classified from different aspects: factoid vs. non-factoid questions, objective vs. subjective/opinion questions, geo-analytical vs. knowledge lookup questions, textual vs. visual questions, and so on.
More specifically, the first five question types are classified based on the types of knowledge that a question focuses on - factoid knowledge, the knowledge about future, the knowledge about people's opinions, common sense knowledge about hypothetical events, or knowledge about the explanations for geographic facts. Geo-analytical questions are listed as one specific type because of its specific focus on GIS workflow synthesis. % which is a unique GeoQA process compared with a normal fact lookup QA process. 
The scenario-based and visual geographic question types emphasize the context (e.g., text description, images)  associated with the question. Basically, these question types reflect different aspects and focuses of GeoQA.}

\revise{These question types are not necessarily mutually exclusive from each other. For example, as for Question \textit{What would be the best location if we want to build a new elementary school in Seattle}, it is both a hypothetical geographic question and a geo-analytical question because this question follows the "what would happen if..." hypothetical question pattern and answering it requires GIS workflow synthesis (e.g., site selection analysis). Question \textit{How many buildings are in the current remote sensing image} is both a factoid geographic question and a visual geographic question. }

\revise{Moreover, this question classification only reflects our current understanding of GeoQA research and is by no means a final and complete system for geographic question classification. With the advancement of the GeoQA research, we might see new types of geographic questions which have not been covered by the presented classification system. %However, we believe the current classification system can guide the development of future GeoQA systems.
}

\revise{Nevertheless, we believe the presented geographic question classification is useful since it can help a GeoQA researcher to narrow down the focus and find an appropriate GeoQA dataset that fits into their research scope. It can also guide them in the process of GeoQA benchmark dataset construction and analysis as \citet{hamzei2019place} did. %It can also help them to find relevant research and pay attention to the difference among different GeoQA work during literature review.
Last but not least , a question classification system helps identify the challenges and future research directions for GeoQA.
}

\section{Future Research Directions for GeoQA}  \label{sec:futuredir}

In this section, we will discuss some interesting research directions for GeoQA. Most importantly, we need to address the question of what unique contributions we can make in GeoQA beyond work on more general AQ systems.

Question answering is one of the most important research topics in natural language processing. Currently, there are around 30 different large-scale question answering data sets available\footnote{\url{http://nlpprogress.com/english/question_answering.html}}. Most of them are about reading comprehension and open-domain question answering such as HotpotQA~\citep{yang2019end}, SQuAD ~\citep{chen2017reading}, Natural Questions~\citep{kwiatkowski2019natural}, CoQA~\citep{reddy2019coqa} \revise{which mainly aim at} unstructured-text based QA. There are also QA datasets for structured-knowledge-based QA such as QALD-9~\citep{ngomo2018}. 
%Every year there are thousands of questions answering papers published in different NLP or machine learning conferences such as  

Compared with QA, GeoQA is a smaller research topic which starts attracting attentions from QA researchers as well as GIScientists only recently. 
\revise{A recent review on the usage of geospatial information in virtual assistants \citep{granellscoping2021} also showed that the usage of different types of geographic data and various spatial methods in virtual assistants is quite limited.}
%Figure \ref{fig:qa-paper} and \ref{fig:geoqa-paper} show the Google search results for papers on arxiv.org about question answering and geographic question answering. There are more than 55,000 papers on arxiv.org about QA while we only get 1560 search results for GeoQA. Note that based on our manual investigation, in the search results for GeoQA, almost all papers after the top 7 results are not really about GeoQA but normal QA papers which just include the term ``geographic'' once or twice in the discussion section. Given the fact that GeoQA starts at the age when QA research is far more developed and mature,
\textit{How we can show the unique contribution of GeoQA to the general QA community} is the golden question needed to be answered for GIScientists. %We need to utilize the existing QA method if necessary and try to avoid reinventing the wheels. 

% \begin{figure*}[!ht]
%     \centering
%     \begin{subfigure}[b]{0.48\textwidth}
%         \centering
%         \includegraphics[width=\textwidth]{./fig/qa-paper.png}
%         \caption[]%
% 		{Google search for papers about QA
% 		}
%         \label{fig:qa-paper}
%     \end{subfigure}
%     \hfill
%     \begin{subfigure}[b]{0.48\textwidth}
%         \centering
%         \includegraphics[width=\textwidth]{./fig/geoqa-paper.png}
%         \caption[]%
% 		{Google search for papers about GeoQA
% 		}
%         \label{fig:geoqa-paper}
%     \end{subfigure}
%         \caption{The screenshots of two Google searches that seek for papers in arxiv.org about \textit{question answering} and \textit{geographic question answering}.}
%     \label{fig:qa-paper-search}
% \end{figure*}

As far as we see, there are some interesting and unique research directions specifically for GeoQA:
\begin{enumerate}
    \item \textbf{How to effectively utilize geographic coordinates in a GeoQA model?} As the basic element of geographic information, how to effectively utilize locations in deep learning models for any geospatial task is a fundamental problem itself. \cite{contractor2020joint} presented an indirect way to encode distances among locations (e.g., POIs) for the GeoQA purpose. In contrast, \citet{mac2019presence,mai2020multi,mai2020se} take a more explicit approach which directly encode coordinates into location embeddings for multiple downstream tasks. Which one works better for a specific GeoQA task needs to be investigated.
    
    \item \textbf{How to effectively utilize complex spatial footprints of geographic entities such as polygons, multipolygons, and polylines in a GeoQA model? How to design efficient ``fuzzy spatial operators'' which are robust to the geometry uncertainty problem?} These complex spatial footprints are essential for many geographic question types. However, as we discussed in Section \ref{sec:geo-uncertain}, directly utilizing deterministic spatial operators such as GeoSPARQL functions as \citet{punjani2018template} did will suffer from the known problems with using raw geometries which will \revise{affect} the performance of GeoQA.  
    %Developing an efficient neural-network-based ``fuzzy spatial operator'' which is robust to geometric uncertainty is a promising research direction.
    A more proper way is to design an efficient neural-network-based ``fuzzy spatial operator'' which is robust to the geometric uncertainty problem.
    This ``fuzzy spatial operator'' takes these complex polygon geometries as input and outputs their spatial relations. At the training phase, this operator automatically learns the concept of thresholds implicitly based on the training labels and we \revise{do} not need to specify thresholds explicitly as \citet{regalia2019computing} did. This might be an interesting research direction.
    
    \item \textbf{How to define a compact but effective set of spatial operators for GeoQA? Furthermore, how to define a program language similar to Lisp \citep{liang2017neural} and Prolog \citep{zelle1996learning} but for spatial computing which will make GeoQA easier?} As we discussed in Section \ref{sec:unique}, given the large number of spatial operators, we need to derive a small subset which can be used to answer most of the geographic question types. The core concepts of spatial information research \citep{kuhn2012core} may be a great starting point since it provides a list of core spatial operators/computations and defines a high-level language for spatial computing
    \citep{kuhn2015designing}. However, several issues need to be investigated further - How good are these spatial operators? How easily can they be applied to GeoQA? And how many question types can they support?
    
    \item \textbf{How to handle the vagueness of spatial relations as well as geographic concepts in a GeoQA model?} 
    \revise{The selection of spatial operators should be aware of the vagueness of geographic concepts and geographic entities during question answering process. For example, 
    %the spatial operator used for answer Question \textit{Find a good house for renting along the State Street} should be different from that for Question \textit{Find a good house for renting along the Mississippi River}. Aside from the map scale problem, the spatial relation \textit{along a street} is well-defined while the spatial relation \textit{along a river} is conceptually vague because of the vagueness of the concept \textit{river}. Similarly, 
    Question \textit{Is San Luis Obispo part of Southern California} and Question \textit{Is San Luis Obispo part of California} should be handled differently. Unlike California, Southern California is a vague cognitive region which does not have a crisp boundary. The ordinary topological relation operators cannot deal with this.
    %The vagueness of geographic concepts and entities is a very complex issue. 
    It might be complicated to design a GeoQA model to directly interpret the vagueness of geographic concepts and entities. }
    %\revise{It is also challenging to write QA rules to handle each individual cases which is a normal practice for many commercial QA systems such as Apple Siri and Google Assistant. }
    A simple yet effective approach is to collect annotated data for QA pairs which contain these spatial operators and concepts and develop an end-to-end model to learn from them. 
\end{enumerate}

In this paper, we attempt to provide a holistic view of the current landscape of GeoQA research as well as its challenges and uniqueness. %As John Adams once said, ``Every problem is an opportunity in disguise.'' 
We hope the GeoQA problem mentioned by Jordan can be solved and a real geospatial artificial intelligence agent can be built in the coming years.

\section{Software and Data Availability}
The data utilized in this paper are downloaded from OpenStreetMap and visualized using QGIS. All data and software used are open source.

\begin{acknowledgements}
\revise{This work is funded by the NSF award 2033521 A1: \textit{KnowWhereGraph: Enriching and Linking Cross-Domain Knowledge Graphs using Spatially-Explicit AI Technologies}.}
\end{acknowledgements}
%\begin{acknowledgements}
%\end{acknowledgements}

\setlength\bibhang{0pt}
\setlength\bibsep{4pt}

% \renewcommand\refname{\textbf{\large References}} 
%\begin{thebibliography}{}

%\bibitem{author1}
%Author, F. and Author, S.: The test article, International Journal of %GIScience., 12, 135–147, \uline{\url{https://doi.org/10.1234/56789}}, 2019.

%\bibitem{author2}
%Author, O., Author, T. and Author, F.: More test articles, Transactions in %GIS., 35, 13–28, \uline{\url{https://doi.org/10.2345/67890}}, 2020.

%\end{thebibliography}
%\bibitem{author2}

%% Since the AGILE LaTeX package includes the BibTeX style file agile.bst,authors experienced with BibTeX only have to include the following two lines:
%%

 \bibliographystyle{agile}
 \bibliography{reference.bib}

\begin{thebibliography}{68}
\providecommand{\natexlab}[1]{#1}
\providecommand{\url}[1]{{\tt #1}}
\providecommand{\urlprefix}{URL }
\expandafter\ifx\csname urlstyle\endcsname\relax
  \providecommand{\doi}[1]{https://doi.org/\discretionary{}{}{}#1}\else
  \providecommand{\doi}{https://doi.org/\discretionary{}{}{}\begingroup
  \urlstyle{rm}\Url}\fi

\bibitem[{Antol et~al.(2015)Antol, Agrawal, Lu, Mitchell, Batra, Zitnick, and
  Parikh}]{antol2015vqa}
Antol, S., Agrawal, A., Lu, J., Mitchell, M., Batra, D., Zitnick, C.~L., and
  Parikh, D.: {VQA}: Visual question answering, in: Proceedings of the IEEE
  international conference on computer vision, pp. 2425--2433, 2015.

\bibitem[{Asai et~al.(2020)Asai, Hashimoto, Hajishirzi, Socher, and
  Xiong}]{asai2020learning}
Asai, A., Hashimoto, K., Hajishirzi, H., Socher, R., and Xiong, C.: Learning to
  Retrieve Reasoning Paths over Wikipedia Graph for Question Answering, in:
  International Conference on Learning Representations, 2020.

\bibitem[{Battle and Kolas(2012)}]{battle2012enabling}
Battle, R. and Kolas, D.: Enabling the geospatial semantic web with parliament
  and geosparql, Semantic Web, 3, 355--370, 2012.

\bibitem[{Bennett(2002)}]{bennett2002forest}
Bennett, B.: What is a forest? On the vagueness of certain geographic concepts,
  in: Topoi, Citeseer, 2002.

\bibitem[{Berant et~al.(2013)Berant, Chou, Frostig, and
  Liang}]{berant2013semantic}
Berant, J., Chou, A., Frostig, R., and Liang, P.: Semantic parsing on freebase
  from question-answer pairs, in: Proceedings of the 2013 conference on
  empirical methods in natural language processing, pp. 1533--1544, 2013.

\bibitem[{Billen and Clementini(2004)}]{billen2004model}
Billen, R. and Clementini, E.: A model for ternary projective relations between
  regions, in: International Conference on Extending Database Technology, pp.
  310--328, Springer, 2004.

\bibitem[{Chen et~al.(2017)Chen, Fisch, Weston, and Bordes}]{chen2017reading}
Chen, D., Fisch, A., Weston, J., and Bordes, A.: Reading Wikipedia to Answer
  Open-Domain Questions, in: Proceedings of the 55th Annual Meeting of the
  Association for Computational Linguistics (Volume 1: Long Papers), vol.~1,
  pp. 1870--1879, 2017.

\bibitem[{Chen(2014)}]{chen2014parameterized}
Chen, W.: Parameterized Spatial SQL Translation for Geographic Question
  Answering, in: 2014 IEEE International Conference on Semantic Computing, pp.
  23--27, IEEE, 2014.

\bibitem[{Chen et~al.(2013)Chen, Fosler-Lussier, Xiao, Raje, Ramnath, and
  Sui}]{chen2013synergistic}
Chen, W., Fosler-Lussier, E., Xiao, N., Raje, S., Ramnath, R., and Sui, D.: A
  synergistic framework for geographic question answering, in: 2013 IEEE
  Seventh International Conference on Semantic Computing, pp. 94--99, IEEE,
  2013.

\bibitem[{Chen et~al.(2020)Chen, Liang, Yu, Zhou, Song, and
  Le}]{chen2020neural}
Chen, X., Liang, C., Yu, A.~W., Zhou, D., Song, D., and Le, Q.~V.: Neural
  symbolic reader: Scalable integration of distributed and symbolic
  representations for reading comprehension, in: International Conference on
  Learning Representations, 2020.

\bibitem[{Cohn et~al.(1997)Cohn, Bennett, Gooday, and
  Gotts}]{cohn1997qualitative}
Cohn, A.~G., Bennett, B., Gooday, J., and Gotts, N.~M.: Qualitative spatial
  representation and reasoning with the region connection calculus,
  GeoInformatica, 1, 275--316, 1997.

\bibitem[{Contractor et~al.(2019)Contractor, Shah, Partap, Singla
  et~al.}]{contractor2019large}
Contractor, D., Shah, K., Partap, A., Singla, P., et~al.: Large scale question
  answering using tourism data, arXiv preprint arXiv:1909.03527, 2019.

\bibitem[{Contractor et~al.(2020)Contractor, Goel, Singla
  et~al.}]{contractor2020joint}
Contractor, D., Goel, S., Singla, P., et~al.: Joint Spatio-Textual Reasoning
  for Answering Tourism Questions, arXiv preprint arXiv:2009.13613, 2020.

\bibitem[{Egenhofer and Dube(2009)}]{egenhofer2009topological}
Egenhofer, M.~J. and Dube, M.~P.: Topological relations from metric
  refinements, in: Proceedings of the 17th ACM SIGSPATIAL International
  Conference on Advances in Geographic Information Systems, pp. 158--167, 2009.

\bibitem[{Frank(1992)}]{frank1992qualitative}
Frank, A.~U.: Qualitative spatial reasoning about distances and directions in
  geographic space, Journal of Visual Languages \& Computing, 3, 343--371,
  1992.

\bibitem[{Gao et~al.(2017)Gao, Janowicz, Montello, Hu, Yang, McKenzie, Ju,
  Gong, Adams, and Yan}]{gao2017data}
Gao, S., Janowicz, K., Montello, D.~R., Hu, Y., Yang, J.-A., McKenzie, G., Ju,
  Y., Gong, L., Adams, B., and Yan, B.: A data-synthesis-driven method for
  detecting and extracting vague cognitive regions, International Journal of
  Geographical Information Science, 31, 1245--1271, 2017.

\bibitem[{Gomes(2014)}]{jordan2014}
Gomes, L.: Machine-Learning Maestro Michael Jordan on the Delusions of Big Data
  and Other Huge Engineering Efforts, 2014.

\bibitem[{Granell et~al.(2021)Granell, Pes{\'a}ntez-Cabrera,
  Vilches-Bl{\'a}zquez, Achig, Luaces, Corti{\~n}as-{\'A}lvarez, Chayle, and
  Morocho}]{granellscoping2021}
Granell, C., Pes{\'a}ntez-Cabrera, P.~G., Vilches-Bl{\'a}zquez, L.~M., Achig,
  R., Luaces, M.~R., Corti{\~n}as-{\'A}lvarez, A., Chayle, C., and Morocho, V.:
  A scoping review on the use, processing and fusion of geographic data in
  virtual assistants, Transactions in GIS, 2021.

\bibitem[{Hamzei et~al.(2019)Hamzei, Li, Vasardani, Baldwin, Winter, and
  Tomko}]{hamzei2019place}
Hamzei, E., Li, H., Vasardani, M., Baldwin, T., Winter, S., and Tomko, M.:
  Place questions and human-generated answers: A data analysis approach, in:
  International Conference on Geographic Information Science, pp. 3--19,
  Springer, 2019.

\bibitem[{Hu et~al.(2015)Hu, Janowicz, Prasad, and Gao}]{hu2015metadata}
Hu, Y., Janowicz, K., Prasad, S., and Gao, S.: Metadata topic harmonization and
  semantic search for linked-data-driven geoportals: A case study using ArcGIS
  Online, Transactions in GIS, 19, 398--416, 2015.

\bibitem[{Huang et~al.(2019)Huang, Shen, Li, Cheng, Zhou, Dai, Qu
  et~al.}]{huang2019geosqa}
Huang, Z., Shen, Y., Li, X., Cheng, G., Zhou, L., Dai, X., Qu, Y., et~al.:
  GeoSQA: A Benchmark for Scenario-based Question Answering in the Geography
  Domain at High School Level, in: Proceedings of the 2019 Conference on
  Empirical Methods in Natural Language Processing and the 9th International
  Joint Conference on Natural Language Processing (EMNLP-IJCNLP), pp.
  5869--5874, 2019.

\bibitem[{Janowicz et~al.(2015)Janowicz, Van~Harmelen, Hendler, and
  Hitzler}]{janowicz2015data}
Janowicz, K., Van~Harmelen, F., Hendler, J.~A., and Hitzler, P.: Why the data
  train needs semantic rails, AI Magazine, 36, 5--14, 2015.

\bibitem[{Janowicz et~al.(2020)Janowicz, Gao, McKenzie, Hu, and
  Bhaduri}]{janowicz2020geoai}
Janowicz, K., Gao, S., McKenzie, G., Hu, Y., and Bhaduri, B.: GeoAI: Spatially
  explicit artificial intelligence techniques for geographic knowledge
  discovery and beyond, International Journal of Geographical Information
  Science, 2020.

\bibitem[{Jiang et~al.(2018)Jiang, Li, Yang, Hu, Armstrong, Huang, Moroni,
  McGibbney, and Finch}]{jiang2018towards}
Jiang, Y., Li, Y., Yang, C., Hu, F., Armstrong, E.~M., Huang, T., Moroni, D.,
  McGibbney, L.~J., and Finch, C.~J.: Towards intelligent geospatial data
  discovery: a machine learning framework for search ranking, International
  journal of digital earth, 11, 956--971, 2018.

\bibitem[{Karpukhin et~al.(2020)Karpukhin, Oguz, Min, Lewis, Wu, Edunov, Chen,
  and Yih}]{karpukhin2020dense}
Karpukhin, V., Oguz, B., Min, S., Lewis, P., Wu, L., Edunov, S., Chen, D., and
  Yih, W.-t.: Dense Passage Retrieval for Open-Domain Question Answering, in:
  Proceedings of the 2020 Conference on Empirical Methods in Natural Language
  Processing (EMNLP), pp. 6769--6781, 2020.

\bibitem[{Khan et~al.(2014)Khan, Baharudin, Khan, and Ullah}]{khan2014mining}
Khan, K., Baharudin, B., Khan, A., and Ullah, A.: Mining opinion components
  from unstructured reviews: A review, Journal of King Saud University-Computer
  and Information Sciences, 26, 258--275, 2014.

\bibitem[{Kordjamshidi et~al.(2020)Kordjamshidi, Pustejovsky, and
  Moens}]{kordjamshidi2020representation}
Kordjamshidi, P., Pustejovsky, J., and Moens, M.~F.: Representation, Learning
  and Reasoning on Spatial Language for Downstream NLP Tasks, in: Proceedings
  of the 2020 Conference on Empirical Methods in Natural Language Processing:
  Tutorial Abstracts, pp. 28--33, 2020.

\bibitem[{Kuhn(2003)}]{kuhn2003semantic}
Kuhn, W.: Semantic reference systems, International Journal of Geographical
  Information Science, 17, 405--409, 2003.

\bibitem[{Kuhn(2012)}]{kuhn2012core}
Kuhn, W.: Core concepts of spatial information for transdisciplinary research,
  International Journal of Geographical Information Science, 26, 2267--2276,
  2012.

\bibitem[{Kuhn and Ballatore(2015)}]{kuhn2015designing}
Kuhn, W. and Ballatore, A.: Designing a language for spatial computing, in:
  AGILE 2015, pp. 309--326, Springer, 2015.

\bibitem[{Kwiatkowski et~al.(2019)Kwiatkowski, Palomaki, Redfield, Collins,
  Parikh, Alberti, Epstein, Polosukhin, Devlin, Lee
  et~al.}]{kwiatkowski2019natural}
Kwiatkowski, T., Palomaki, J., Redfield, O., Collins, M., Parikh, A., Alberti,
  C., Epstein, D., Polosukhin, I., Devlin, J., Lee, K., et~al.: Natural
  questions: a benchmark for question answering research, Transactions of the
  Association for Computational Linguistics, 7, 453--466, 2019.

\bibitem[{Liang et~al.(2017)Liang, Berant, Le, Forbus, and
  Lao}]{liang2017neural}
Liang, C., Berant, J., Le, Q., Forbus, K.~D., and Lao, N.: Neural Symbolic
  Machines: Learning Semantic Parsers on Freebase with Weak Supervision, in:
  Proceedings of the 55th Annual Meeting of the Association for Computational
  Linguistics (Volume 1: Long Papers), vol.~1, pp. 23--33, 2017.

\bibitem[{Liang et~al.(2018)Liang, Norouzi, Berant, Le, and
  Lao}]{liang2018memory}
Liang, C., Norouzi, M., Berant, J., Le, Q.~V., and Lao, N.: Memory Augmented
  Policy Optimization for Program Synthesis and Semantic Parsing, in: Advances
  in Neural Information Processing Systems, pp. 10\,014--10\,026, 2018.

\bibitem[{Lobry et~al.(2020)Lobry, Marcos, Murray, and Tuia}]{lobry2020rsvqa}
Lobry, S., Marcos, D., Murray, J., and Tuia, D.: RSVQA: Visual question
  answering for remote sensing data, IEEE Transactions on Geoscience and Remote
  Sensing, 58, 8555--8566, 2020.

\bibitem[{Mac~Aodha et~al.(2019)Mac~Aodha, Cole, and Perona}]{mac2019presence}
Mac~Aodha, O., Cole, E., and Perona, P.: Presence-only geographical priors for
  fine-grained image classification, in: Proceedings of the IEEE/CVF
  International Conference on Computer Vision, pp. 9596--9606, 2019.

\bibitem[{Mai et~al.(2018)Mai, Janowicz, He, Liu, and Lao}]{mai2018poireviewqa}
Mai, G., Janowicz, K., He, C., Liu, S., and Lao, N.: {POIR}eview{QA}: A
  Semantically Enriched {POI} Retrieval and Question Answering Dataset, in:
  Proceedings of the 12th Workshop on Geographic Information Retrieval, p.~5,
  ACM, 2018.

\bibitem[{Mai et~al.(2019)Mai, Yan, Janowicz, and Zhu}]{mai2019relaxing}
Mai, G., Yan, B., Janowicz, K., and Zhu, R.: Relaxing unanswerable geographic
  questions using a spatially explicit knowledge graph embedding model, in:
  AGILE, pp. 21--39, Springer, 2019.

\bibitem[{Mai et~al.(2020{\natexlab{a}})Mai, Janowicz, Cai, Zhu, Regalia, Yan,
  Shi, and Lao}]{mai2020se}
Mai, G., Janowicz, K., Cai, L., Zhu, R., Regalia, B., Yan, B., Shi, M., and
  Lao, N.: SE-KGE: A location-aware Knowledge Graph Embedding model for
  Geographic Question Answering and Spatial Semantic Lifting, Transactions in
  GIS, 24, 623--655, 2020{\natexlab{a}}.

\bibitem[{Mai et~al.(2020{\natexlab{b}})Mai, Janowicz, Prasad, Shi, Cai, Zhu,
  Regalia, and Lao}]{mai2020semantically}
Mai, G., Janowicz, K., Prasad, S., Shi, M., Cai, L., Zhu, R., Regalia, B., and
  Lao, N.: Semantically-Enriched Search Engine for Geoportals: A Case Study
  with ArcGIS Online, AGILE: GIScience Series, 1, 1--17, 2020{\natexlab{b}}.

\bibitem[{Mai et~al.(2020{\natexlab{c}})Mai, Janowicz, Yan, Zhu, Cai, and
  Lao}]{mai2020multi}
Mai, G., Janowicz, K., Yan, B., Zhu, R., Cai, L., and Lao, N.: Multi-Scale
  Representation Learning for Spatial Feature Distributions using Grid Cells,
  in: International Conference on Learning Representations, 2020{\natexlab{c}}.

\bibitem[{Miller et~al.(2016)Miller, Fisch, Dodge, Karimi, Bordes, and
  Weston}]{miller2016key}
Miller, A., Fisch, A., Dodge, J., Karimi, A.-H., Bordes, A., and Weston, J.:
  Key-value memory networks for directly reading documents, in: Empirical
  Methods in Natural Language Processing (EMNLP), pp. 1400--1409, 2016.

\bibitem[{Mishra and Jain(2016)}]{mishra2016survey}
Mishra, A. and Jain, S.~K.: A survey on question answering systems with
  classification, Journal of King Saud University-Computer and Information
  Sciences, 28, 345--361, 2016.

\bibitem[{Montello et~al.(2003)Montello, Goodchild, Gottsegen, and
  Fohl}]{montello2003s}
Montello, D.~R., Goodchild, M.~F., Gottsegen, J., and Fohl, P.: Where's
  downtown?: Behavioral methods for determining referents of vague spatial
  queries, Spatial Cognition \& Computation, 3, 185--204, 2003.

\bibitem[{Montello et~al.(2014)Montello, Friedman, and
  Phillips}]{montello2014vague}
Montello, D.~R., Friedman, A., and Phillips, D.~W.: Vague cognitive regions in
  geography and geographic information science, International Journal of
  Geographical Information Science, 28, 1802--1820, 2014.

\bibitem[{Moretti(1998)}]{moretti1998atlas}
Moretti, F.: Atlas of the European novel, 1800-1900, London: Verso, 1998.

\bibitem[{Ngomo(2018)}]{ngomo2018}
Ngomo, N.: 9th Challenge on Question Answering over Linked Data (QALD-9),
  language, 7, 1, 2018.

\bibitem[{Pasupat and Liang(2015)}]{pasupat2015compositional}
Pasupat, P. and Liang, P.: Compositional Semantic Parsing on Semi-Structured
  Tables, in: Proceedings of the 53rd Annual Meeting of the Association for
  Computational Linguistics and the 7th International Joint Conference on
  Natural Language Processing (Volume 1: Long Papers), vol.~1, pp. 1470--1480,
  2015.

\bibitem[{Punjani et~al.(2018)Punjani, Singh, Both, Koubarakis, Angelidis,
  Bereta, Beris, Bilidas, Ioannidis, Karalis et~al.}]{punjani2018template}
Punjani, D., Singh, K., Both, A., Koubarakis, M., Angelidis, I., Bereta, K.,
  Beris, T., Bilidas, D., Ioannidis, T., Karalis, N., et~al.: Template-based
  question answering over linked geospatial data, in: Proceedings of the 12th
  Workshop on Geographic Information Retrieval, pp. 1--10, 2018.

\bibitem[{Rajpurkar et~al.(2016)Rajpurkar, Zhang, Lopyrev, and
  Liang}]{rajpurkar2016squad}
Rajpurkar, P., Zhang, J., Lopyrev, K., and Liang, P.: {SQ}u{AD}: 100,000+
  Questions for Machine Comprehension of Text, in: Proceedings of the 2016
  Conference on Empirical Methods in Natural Language Processing, pp.
  2383--2392, 2016.

\bibitem[{Ramalho et~al.(2018)Ramalho, Ko{\v{c}}isk{\`y}, Besse, Eslami, Melis,
  Viola, Blunsom, and Hermann}]{ramalho2018encoding}
Ramalho, T., Ko{\v{c}}isk{\`y}, T., Besse, F., Eslami, S., Melis, G., Viola,
  F., Blunsom, P., and Hermann, K.~M.: Encoding spatial relations from natural
  language, arXiv preprint arXiv:1807.01670, 2018.

\bibitem[{Reddy et~al.(2019)Reddy, Chen, and Manning}]{reddy2019coqa}
Reddy, S., Chen, D., and Manning, C.~D.: Coqa: A conversational question
  answering challenge, Transactions of the Association for Computational
  Linguistics, 7, 249--266, 2019.

\bibitem[{Regalia et~al.(2016)Regalia, Janowicz, and Gao}]{regalia2016volt}
Regalia, B., Janowicz, K., and Gao, S.: VOLT: a provenance-producing,
  transparent SPARQL proxy for the on-demand computation of linked data and its
  application to spatiotemporally dependent data, in: European Semantic Web
  Conference, pp. 523--538, Springer, 2016.

\bibitem[{Regalia et~al.(2017)Regalia, Janowicz, and
  McKenzie}]{regalia2017revisiting}
Regalia, B., Janowicz, K., and McKenzie, G.: Revisiting the representation of
  and need for raw geometries on the linked data web, in: LDOW@ WWW, 2017.

\bibitem[{Regalia et~al.(2019)Regalia, Janowicz, and
  McKenzie}]{regalia2019computing}
Regalia, B., Janowicz, K., and McKenzie, G.: Computing and querying strict,
  approximate, and metrically refined topological relations in linked
  geographic data, Transactions in GIS, 23, 601--619, 2019.

\bibitem[{Scheider et~al.(2019)Scheider, Ballatore, and
  Lemmens}]{scheider2019finding}
Scheider, S., Ballatore, A., and Lemmens, R.: Finding and sharing GIS methods
  based on the questions they answer, International journal of digital earth,
  12, 594--613, 2019.

\bibitem[{Scheider et~al.(2020)Scheider, Nyamsuren, Kruiger, and
  Xu}]{scheider2020geo}
Scheider, S., Nyamsuren, E., Kruiger, H., and Xu, H.: Geo-analytical
  question-answering with GIS, International Journal of Digital Earth, pp.
  1--14, 2020.

\bibitem[{Sun et~al.(2018)Sun, Dhingra, Zaheer, Mazaitis, Salakhutdinov, and
  Cohen}]{sun2018open}
Sun, H., Dhingra, B., Zaheer, M., Mazaitis, K., Salakhutdinov, R., and Cohen,
  W.: Open Domain Question Answering Using Early Fusion of Knowledge Bases and
  Text, in: Proceedings of the 2018 Conference on Empirical Methods in Natural
  Language Processing, pp. 4231--4242, 2018.

\bibitem[{Turing(1950)}]{turing1950}
Turing, A.~M.: Computing Machinery and Intelligence, Mind, 59, 433--460, 1950.

\bibitem[{Worboys(2001)}]{worboys2001nearness}
Worboys, M.~F.: Nearness relations in environmental space, International
  Journal of Geographical Information Science, 15, 633--651, 2001.

\bibitem[{Xiong et~al.(2019)Xiong, Yu, Chang, Guo, and
  Wang}]{xiong2019improving}
Xiong, W., Yu, M., Chang, S., Guo, X., and Wang, W.~Y.: Improving Question
  Answering over Incomplete KBs with Knowledge-Aware Reader, in: Proceedings of
  the 57th Annual Meeting of the Association for Computational Linguistics, pp.
  4258--4264, 2019.

\bibitem[{Xiong et~al.(2020)Xiong, Li, Iyer, Du, Lewis, Wang, Mehdad, Yih,
  Riedel, Kiela et~al.}]{xiong2020answering}
Xiong, W., Li, X.~L., Iyer, S., Du, J., Lewis, P., Wang, W.~Y., Mehdad, Y.,
  Yih, W.-t., Riedel, S., Kiela, D., et~al.: Answering Complex Open-Domain
  Questions with Multi-Hop Dense Retrieval, arXiv preprint arXiv:2009.12756,
  2020.

\bibitem[{Xu et~al.(2020)Xu, Hamzei, Nyamsuren, Kruiger, Winter, Tomko, and
  Scheider}]{xu2020extracting}
Xu, H., Hamzei, E., Nyamsuren, E., Kruiger, H., Winter, S., Tomko, M., and
  Scheider, S.: Extracting interrogative intents and concepts from geo-analytic
  questions, AGILE: GIScience Series, 1, 1--21, 2020.

\bibitem[{Yang et~al.(2017)Yang, Nie, Cohen, and Lao}]{yang2017learning}
Yang, F., Nie, J., Cohen, W.~W., and Lao, N.: Learning to Organize Knowledge
  with N-Gram Machines, arXiv preprint arXiv:1711.06744, 2017.

\bibitem[{Yang et~al.(2019)Yang, Xie, Lin, Li, Tan, Xiong, Li, and
  Lin}]{yang2019end}
Yang, W., Xie, Y., Lin, A., Li, X., Tan, L., Xiong, K., Li, M., and Lin, J.:
  End-to-End Open-Domain Question Answering with BERTserini, in: Proceedings of
  the 2019 Conference of the North American Chapter of the Association for
  Computational Linguistics (Demonstrations), pp. 72--77, 2019.

\bibitem[{Yih et~al.(2015)Yih, Chang, He, and Gao}]{yih2015semantic}
Yih, W.-t., Chang, M.-W., He, X., and Gao, J.: Semantic Parsing via Staged
  Query Graph Generation: Question Answering with Knowledge Base, in:
  Proceedings of the 53rd Annual Meeting of the Association for Computational
  Linguistics and the 7th International Joint Conference on Natural Language
  Processing (Volume 1: Long Papers), pp. 1321--1331, 2015.

\bibitem[{Yih et~al.(2016)Yih, Richardson, Meek, Chang, and Suh}]{yih2016value}
Yih, W.-t., Richardson, M., Meek, C., Chang, M.-W., and Suh, J.: The value of
  semantic parse labeling for knowledge base question answering, in:
  Proceedings of the 54th Annual Meeting of the Association for Computational
  Linguistics (Volume 2: Short Papers), vol.~2, pp. 201--206, 2016.

\bibitem[{Yin et~al.(2016)Yin, Yu, Xiang, Zhou, and
  Sch{\"u}tze}]{yin2016simple}
Yin, W., Yu, M., Xiang, B., Zhou, B., and Sch{\"u}tze, H.: Simple Question
  Answering by Attentive Convolutional Neural Network, in: Proceedings of
  COLING 2016, the 26th International Conference on Computational Linguistics:
  Technical Papers, pp. 1746--1756, 2016.

\bibitem[{Zelle and Mooney(1996)}]{zelle1996learning}
Zelle, J.~M. and Mooney, R.~J.: Learning to parse database queries using
  inductive logic programming, in: Proceedings of the national conference on
  artificial intelligence, pp. 1050--1055, 1996.

\end{thebibliography}

\end{document}